\documentclass[10pt,twocolumn,letterpaper]{article}

\usepackage{cvpr}
\usepackage{times}
\usepackage{epsfig}
\usepackage{graphicx}
\usepackage{amsmath}
\usepackage{amssymb}

\usepackage{subfig}
\usepackage{multirow}
\usepackage{rotating}
\usepackage{algorithm}
\usepackage{algpseudocode} %
\usepackage[toc,page]{appendix}

\usepackage[pagebackref=true,breaklinks=true,letterpaper=true,colorlinks,bookmarks=false]{hyperref}

\cvprfinalcopy

\begin{document}

\title{Single-Image Piece-wise Planar 3D Reconstruction via Associative Embedding}

\author{Zehao Yu$^1$\thanks{Equal contribution}\\
\and
Jia Zheng$^{1*}$\\
\and
Dongze Lian$^1$\\
\and
Zihan Zhou$^2$\\
\and
Shenghua Gao$^1$\thanks{Corresponding author}
\and
$^1$ShanghaiTech University\\
{\tt\small \{yuzh,zhengjia,liandz,gaoshh\}@shanghaitech.edu.cn}
\and
$^2$The Pennsylvania State University\\
{\tt\small zzhou@ist.psu.edu}}

\newcommand{\TODO}[1]{{\bf \textcolor{red}{[TODO: #1]}}}
\newcommand{\PAR}[1]{\smallskip \noindent {\bf #1~}}
\newcommand{\PARbegin}[1]{\noindent {\bf #1~}}

\maketitle

\begin{abstract}
Single-image piece-wise planar 3D reconstruction aims to simultaneously segment plane instances and recover 3D plane parameters from an image. Most recent approaches leverage convolutional neural networks (CNNs) and achieve promising results. However, these methods are limited to detecting a fixed number of planes with certain learned order. To tackle this problem, we propose a novel two-stage method based on associative embedding, inspired by its recent success in instance segmentation. In the first stage, we train a CNN to map each pixel to an embedding space where pixels from the same plane instance have similar embeddings. Then, the plane instances are obtained by grouping the embedding vectors in planar regions via an efficient mean shift clustering algorithm. In the second stage, we estimate the parameter for each plane instance by considering both pixel-level and instance-level consistencies. With the proposed method, we are able to detect an arbitrary number of planes. Extensive experiments on public datasets validate the effectiveness and efficiency of our method. Furthermore, our method runs at 30 fps at the testing time, thus could facilitate many real-time applications such as visual SLAM and human-robot interaction. Code is available at \url{https://github.com/svip-lab/PlanarReconstruction}.
\end{abstract}

\section{Introduction}

\begin{figure}[t]
	\centering
	\setlength{\tabcolsep}{0.1em}
	\hfill{}\hspace*{-0.5em}
	\begin{tabular}{cc}
		\includegraphics[width=0.5\columnwidth]{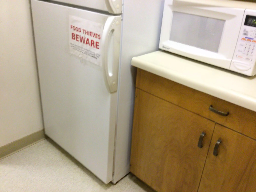}
		&\includegraphics[width=0.5\columnwidth]{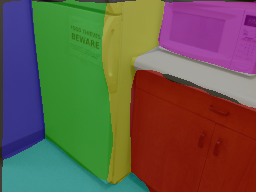}
		\tabularnewline
		{\small{}Input image} &{\small{}Plane instance segmentation}
		\tabularnewline
		\includegraphics[width=0.5\columnwidth]{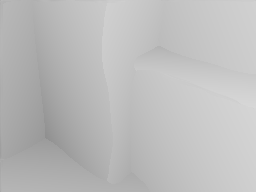} 
		&\includegraphics[width=0.5\columnwidth]{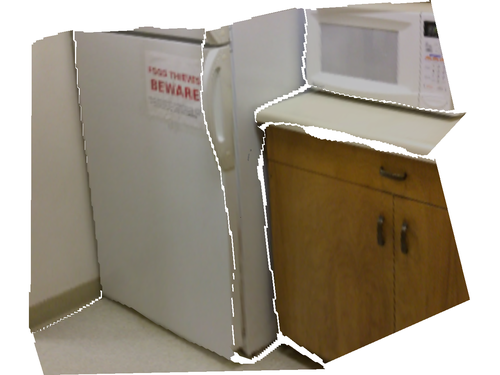}
		\tabularnewline
		{\small{}Depth map} &{\small{}Piece-wise planar 3D model}
		\tabularnewline
	\end{tabular}\hfill{}
	\vspace{-2mm}
	\caption{Piece-wise planar 3D reconstruction.\label{fig:task}}
	\vspace{-4mm}
\end{figure}

Single-image 3D reconstruction is a fundamental problem in computer vision, with many applications in emerging domains such as virtual and augmented reality, robotics, and social media. In this paper, we address this challenging problem by recovering a piece-wise planar 3D model of a scene, that is, to find all the plane instances in a single RGB image and estimate their 3D parameters, as shown in Figure~\ref{fig:task}. The piece-wise planar model provides a compact representation of the 3D scene, which could benefit many applications such as SLAM and human-robot interaction. 

In the literature, most existing methods tackle this problem in a \emph{bottom-up} manner~\cite{Delage2005Automatic,Barinova2008Fast,Micusik2008Towards,Yang2016Efficient,Hoiem2007Recovering,Fouhey2014Unfolding,Haines2015Recognising}. They first extract geometric primitives such as straight line segments, vanishing points, corners, junctions, and image patches from the image. These primitives are then grouped into planar regions based on their geometric relationships. However, in practice, detecting the constituent geometric primitives itself is highly challenging, often resulting in a large number of missed detections (\eg, due to poorly textured surfaces, lighting conditions) and outliers (\eg, due to the presence of non-planar objects). As a result, statistical techniques such as RANSAC or Markov Random Field (MRF) are commonly employed to produce the final 3D models. But such techniques often break down when the percentage of missed and irrelevant detections is high, and are only applicable to restrictive scenarios (\eg, Manhattan world scenes). Further, the optimization of the statistical model is time-consuming, which greatly limits their application in real-time tasks. 

Different from bottom-up methods, a \emph{top-down} approach~\cite{Han2005Bottom} overcomes the aforementioned difficulties by analyzing the image in a holistic fashion, without resorting to local geometric primitives. Recently, \cite{Liu2018PlaneNet,Yang2018Recovering} train CNNs to directly predict plane segmentation and plane parameters from a single image. These methods are shown to achieve the state-of-the-art performance on multiple indoor and outdoor datasets. Despite their advantages, current learning-based methods come with their own limitations. In particular, due to the lack of prior knowledge about the number and specific order of planes in an image, they are limited to detecting a fixed number of planes with certain learned order, thus may be not flexible enough to handle variations in real-world scene structure. 

In this paper, we propose a novel \emph{CNN-based, bottom-up} approach which takes the best of both worlds, while avoiding the limitations of existing methods. To make this possible, our key insight is that we can detect plane instances in an image by computing the likelihood that two pixels belong to the same plane instance and then use these likelihoods to group similar pixels together. Unlike traditional bottom-up methods which perform grouping on geometric primitives, our similarity metric is based on a deep embedding model, following its recent success in pose estimation~\cite{Newell2017Associative}, object detection~\cite{Law2018CornerNet}, and instance segmentation~\cite{Fathi2017Semantic,Brabandere2017Semantic,Kong2018Recurrent}. Next, we mask the non-planar pixels with a planar/non-planar segmentation map generated by another CNN branch. Finally, an efficient mean shift clustering algorithm is employed to cluster the masked pixel embeddings into plane instances. 

Following the plane instance segmentation, we design a plane parameter network by considering both pixel-level accuracy and instance-level consistencies. We first predict the plane parameter at each pixel, then combine those predictions with the plane instances to generate the parameter of each plane. Note that, unlike existing CNN methods, we restrict our networks to make local predictions (\ie, pixel-wise embedding vectors, and plane parameters) and group these predictions in a bottom-up fashion. This enables our method to generate an arbitrary number of planes and avoid being restricted to any specific order or spatial layout. 

In summary, {\bf our contributions} are as follows: i) We present a novel two-stage deep learning framework for piece-wise planar 3D reconstruction. Based on the deep associate embedding model, we design a multi-branch, end-to-end trainable network which can detect an arbitrary number of planes and estimate their parameters simultaneously. ii) We propose a fast variant of mean shift clustering algorithm to group pixel embeddings into plane instances, which achieves real-time performance at the testing time. iii) Extensive experiments on two challenging datasets, ScanNet~\cite{Dai2017ScanNet} and NYUv2~\cite{Silberman2012Indoor}, validate the effectiveness and efficiency of our method. 

\section{Related Work}

\subsection{Single-View Planar Reconstruction}

\PARbegin{Geometry-based methods.}
Geometry-based methods~\cite{Delage2005Automatic,Barinova2008Fast,Micusik2008Towards,Lee2009Geometric} recover 3D information based on geometric cues in 2D image. For example, Delage~\etal~\cite{Delage2005Automatic} first extract line segments, vanishing points, and superpixels from the image. Then an MRF model is used to label the superpixels with a predefined set of plane classes (\ie, three dominant plane orientations under the Manhattan world assumption). Similarly, Barinova~\etal~\cite{Barinova2008Fast} assume that the environment is composed of a flat ground and vertical walls, and use a Conditional Random Field (CRF) model to label the detected primitives. Lee~\etal~\cite{Lee2009Geometric} detect a collection of line segments and vanishing points in an image, and search for the building model in a hypothesis set that best matches the collection of geometric primitives. However, all these approaches rely on strong assumptions about the scene, which limit their applicability in practice. 

\PAR{Appearance-based methods.}
Appearance-based methods infer geometric properties of an image based on its appearance. Early works~\cite{Hoiem2007Recovering,Fouhey2014Unfolding,Haines2015Recognising} take a \textit{bottom-up} approach. They first predict the orientations of local image patches, and then group the patches with similar orientations to form planar regions. Hoiem~\etal~\cite{Hoiem2007Recovering} define a set of discrete surface layout labels, such as ``support'', ``vertical'', and ``sky'', and use a number of hand-crafted local image features (\eg, color, texture, location, and perspective) to train a model to label each superpixel in an image. Haines and Calway~\cite{Haines2015Recognising} learn to predict continuous 3D orientations for pre-segmented regions and cast plane detection as an optimization problem with an MRF model. Fouhey~\etal~\cite{Fouhey2014Unfolding} first detect convex/concave edges, occlusion boundaries, superpixels, and their orientations, then formulate the grouping problem as a binary quadratic program under the Manhattan world assumption. Our method also falls into this category. Different from existing methods, we cast plane detection as an instance segmentation problem, in which we learn a similarity metric to directly segment plane instances in an image, and then estimate plane parameter for each plane instance. 

\begin{figure*}[t]
	\centering
	\includegraphics[width=0.9\textwidth]{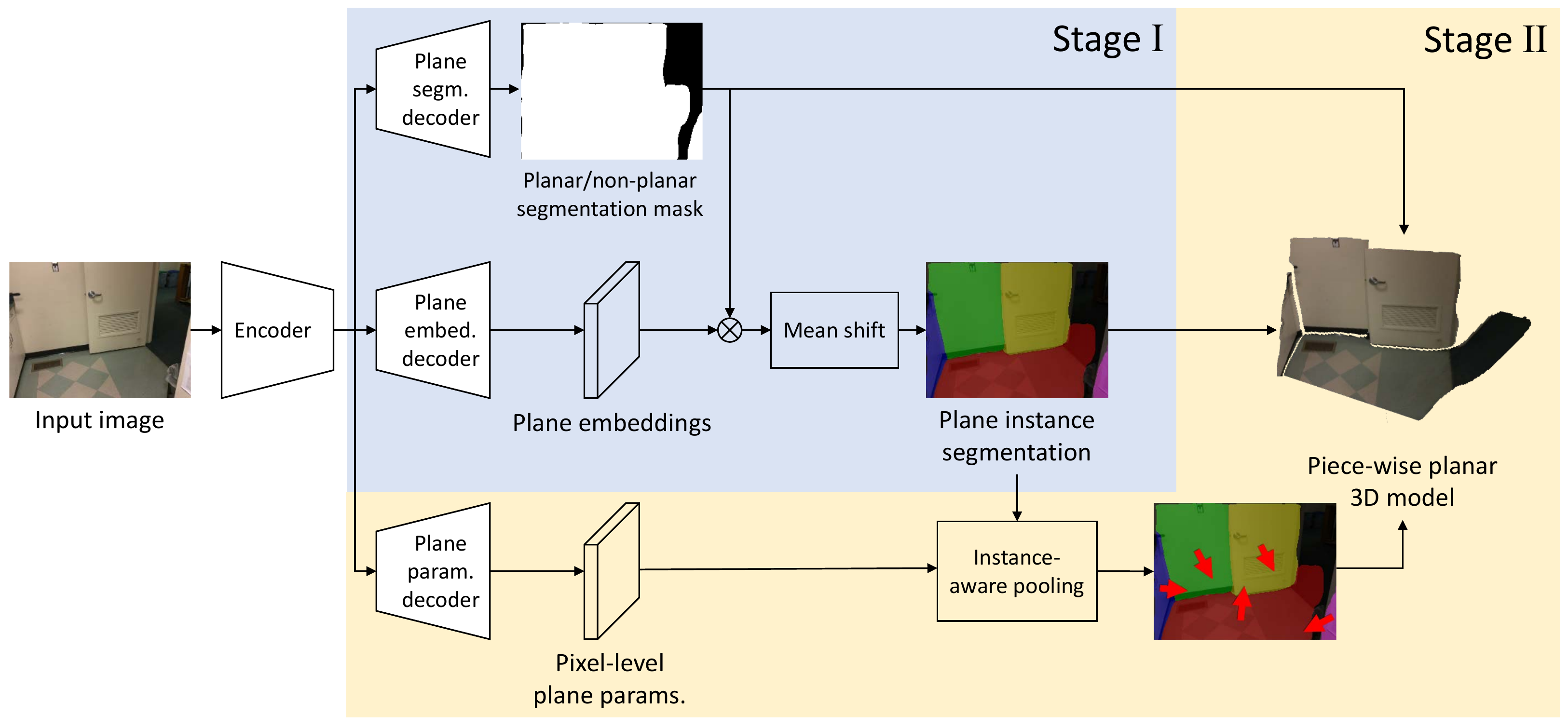}
	\caption{Network architecture. In the {\bf first stage}, the network takes a single RGB image as input, and predicts a planar/non-planar segmentation mask and pixel-level embeddings. Then, an efficient mean shift clustering algorithm is applied to generate plane instances. In the {\bf second stage}, we estimate parameter of each plane by considering both pixel-level and instance-level geometric consistencies.\label{fig:pipeline}}
	\vspace{-2mm}
\end{figure*}

Recently, several CNN-based methods have been proposed to directly predict global 3D plane structures. Liu~\etal~\cite{Liu2018PlaneNet} propose a deep neural network that learns to infer plane parameters and assign plane IDs (segmentation masks) to each pixel in a single image. Yang and Zhou~\cite{Yang2018Recovering} cast the problem as a depth prediction problem and propose a training scheme which does not require ground truth 3D planes. However, these approaches are limited to predicting a fixed number of planes, which could lead to a degraded performance in complex scenes. Concurrently, Liu~\etal~\cite{Liu2018PlaneRCNN} address this problem using a proposal-based instance segmentation framework, \ie, Mask R-CNN~\cite{He2017Mask}. Instead, we leverage a proposal-free instance segmentation approach~\cite{Brabandere2017Semantic} to solve this problem. 

\subsection{Instance Segmentation}
Popular approaches to instance segmentation first generate region proposals, then classify the objects in the bounding box and segment the foreground objects within each proposal~\cite{He2017Mask}. Recent work on associative embedding~\cite{Newell2017Associative} and their extensions in object detection~\cite{Law2018CornerNet} and instance segmentation~\cite{Fathi2017Semantic,Brabandere2017Semantic,Kong2018Recurrent} provide a different solution. These methods learn an embedding function that maps pixels into an embedding space where pixels belonging to the same instance have similar embeddings. Then, they use a simple cluster technique to generate instance segmentation results. Newell~\etal~\cite{Newell2017Associative} introduce associative embedding in the context of multi-person pose estimation and extend it to proposal-free instance segmentation. De~Brabandere~\etal~\cite{Brabandere2017Semantic} propose a discriminative loss to learn the instance embedding, then group embeddings to form instances using a mean shift clustering algorithm. Kong and Fowlkes~\cite{Kong2018Recurrent} introduce a recurrent model to solve the pixel-level clustering problem. Our method is particularly inspired by these work where we treat each plane in an image as an instance, and utilize the idea of associative embedding to detect plane instances. But we further propose i) an efficient mean shift algorithm to cluster plane instances, and ii) an end-to-end trainable network to jointly predict plane instance segmentation and plane parameters, which is not obvious in the context of original instance segmentation problem. 

\section{Method}

Our goal is to infer plane instances and plane parameters from a single RGB image. We propose a novel two-stage method with a multi-branch network to tackle this problem. In the first stage, we train a CNN to obtain planar/non-planar segmentation map and pixel embeddings. We then mask the pixel embeddings with the segmentation map and group the masked pixel embeddings by an efficient mean shift clustering algorithm to form plane instances. In the second stage, we train a network branch to predict pixel-level plane parameters. We then use an instance-aware pooling layer with the instance segmentation map from the first stage to produce the final plane parameters. Figure~\ref{fig:pipeline} shows the overall pipeline of our method. 

\subsection{Planar/Non-Planar Segmentation}
We first design an encoder-decoder architecture to distinguish the planar and non-planar regions. We use an extended version of ResNet-101-FPN~\cite{Lin2017Feature} as an encoder.\footnote{See appendix for more details of network architecture.} The ResNet-101 implemented by \cite{Zhou2017Scene,Zhou2018Semantic} is pretrained on ImageNet~\cite{Deng2009ImageNet} for image classification. The decoder predicts planar/non-planar segmentation map for each pixel. Since the two classes are imbalanced in man-made environments, we use the balanced cross entropy loss as adopted in~\cite{Xie2015Holistically,Caelles2017One}:
\begin{equation}
	L_S = - (1 - w) \sum_{i \in \mathcal{F}} \log p_i - w \sum_{i \in \mathcal{B}} \log (1 - p_i),
\end{equation}
where $\mathcal{F}$ and $\mathcal{B}$ are the set of foreground and background pixels, respectively. $p_i$ is the probability that $i$-th pixel belongs to foreground (\ie, planar regions), and $w$ is the foreground/background pixel-number ratio. 

\subsection{Embedding Model}
Our plane instance segmentation is inspired by recent work on associative embedding~\cite{Newell2017Associative,Fathi2017Semantic,Brabandere2017Semantic,Kong2018Recurrent}. The main idea of associative embedding is to predict an embedding vector for each visual unit such that if some visual units belong to the same instance label, the distance between their embedding vectors should be small so that they can be easily grouped together. 
\begin{figure}[t]
	\centering
	\setlength{\tabcolsep}{0.1em}
	\renewcommand{\arraystretch}{0.1}
	\hfill{}\hspace*{-0.5em}
	\begin{tabular}{cc}
		\includegraphics[width=0.5\columnwidth]{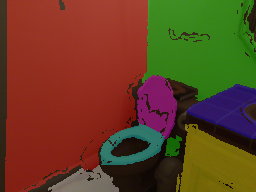} & \includegraphics[width=0.5\columnwidth]{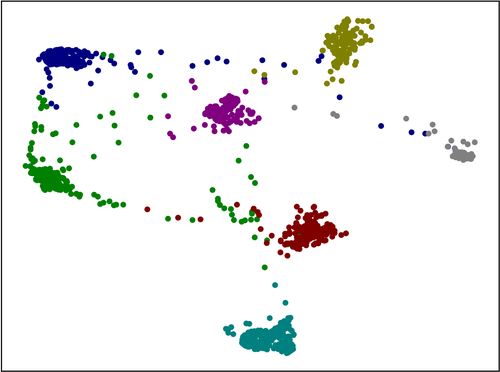}
	\end{tabular}\hfill{}
	\caption{The distribution of plane embeddings. The points with different colors denote learnt embeddings from different plane instances.\label{fig:embedding}}
	\vspace{-2mm}
\end{figure}

For our task, we use a plane embedding branch to map pixels to some embedding space, as shown in Figure~\ref{fig:embedding}. This branch shares the same high-level feature maps with the plane segmentation branch. To enforce pixels in the same plane instance are closer than those in different planes, we use the discriminative loss in \cite{Brabandere2017Semantic}. The loss consists of two terms, namely a ``pull'' loss and a ``push'' loss. The ``pull'' loss pulls each embedding to the mean embedding of the corresponding instance (\ie the instance center), whereas the ``push'' loss pushes the instance centers away from each other.
\begin{equation}
	L_E = L_{pull} + L_{push},
\end{equation}
where
\begin{equation}
	L_{pull} = \frac{1}{C} \sum_{c=1}^{C} \frac{1}{N_c} \sum_{i=1}^{N_c} \max \left( \lVert \mu_c - x_i \rVert - \delta_{\textrm{v}}, 0 \right),
\end{equation}
\begin{equation}
	L_{push} = \frac{1}{C (C-1)} \mathop{\sum_{c_A = 1}^{C} \sum_{c_B = 1}^{C}}_{c_A \neq c_B} \max \left( \delta_{\textrm{d}} - \lVert \mu_{c_A} - \mu_{c_B} \rVert, 0 \right).
\end{equation}
Here, $C$ is the number of clusters $\mathcal{C}$ (planes) in the ground truth, $N_c$ is the number of elements in cluster $c$, $x_i$ is the pixel embedding, $\mu_c$ is the mean embedding of the cluster $c$, and $\delta_{\textrm{v}}$ and $\delta_{\textrm{d}}$ are the margin for ``pull'' and ``push'' losses, respectively.

Intuitively, if the pixel embeddings are easily separable (\ie, the inter-instance distance is larger then $\delta_{\textrm{d}}$, or the distance between an embedding vector and its center is smaller than $\delta_{\textrm{v}}$), the penalty is zero. Otherwise, the penalty will increase sharply. Thus, the loss acts like hard example mining since it only penalizes difficult cases in the embedding. 

\subsection{Efficient Mean Shift Clustering}

Once we have the embedding vector for each pixel, we group them to form plane instances. Mean shift clustering is suitable for this task since the number of plane instances is not known \emph{a priori}. However, the standard mean shift clustering algorithm computes pairwise distance on all pairs of pixel embedding vectors at each iteration. The complexity of each iteration is $O(N^2)$ where $N$ is the number of pixels in the image. In practice, $N$ is very large even for a small size image. For example, in our experiments, $N = 192 \times 256$, making the standard algorithm inapplicable. 

To tackle this problem, we propose a fast variant of the mean shift clustering algorithm. Instead of shifting all pixels in embedding space, we only shift a small number of anchors in embedding space and assign each pixel to the nearest anchor. Specifically, let $k, d$ denote the number of anchors per dimension and the embedding dimension, respectively, we generate $k^d$ anchors uniformly in the embedding space. We then compute pairwise potential between anchor $a_j$ and embedding vector $x_i$ as follows:
\begin{equation}
	p_{ij} =  \frac{1}{\sqrt{2 \pi} b } \exp \left( -\frac{m_{ij}^2}{2 b^2} \right),
	\label{eq:pairwise_potential}
\end{equation}
where $b$ is the bandwidth in mean shift clustering algorithm and $m_{ij} = \|a_j - x_i\|_2$ is the distance between $a_j$ and $x_i$. The shift step of each anchor in each iteration $t$ can be expressed as:
\begin{equation}
	a_j^t = \frac{1}{Z_{j}^t} \sum_{i=1}^N p_{ij}^t \cdot x_i,
	\label{eq:mean_shift}
\end{equation}
where $Z_{j}^t = \sum_{i=1}^N p_{ij}^t$ is a normalization constant. To further speed up the process, we filter out those anchors with low local density at the beginning of clustering. 

After the algorithm converges, we merge nearby anchors to form clusters $\tilde{\mathcal{C}}$, where each cluster $\tilde{c}$ corresponds to a plane instance. Specifically, we consider two anchors belongs to the same cluster if their distance less than bandwidth $b$. The center of this cluster is the mean of anchors belonging to this cluster. 

Finally, we associate pixel embeddings to clusters using soft assignment: 
\begin{equation}
	S_{ij} = \frac{ \exp \left( -m_{ij} \right) } { \sum_{j=1}^{\tilde{C}} \exp \left( -m_{ij} \right)}.
	\label{eq:soft_assignment}
\end{equation}

The details of the proposed algorithm are shown in Algorithm~\ref{alg:mean_shift}. Note that the bandwidth $b$ can be determined by the desired margin in the training stage of vector embedding. The complexity of each iteration of our algorithm is $O(k^d N)$. As long as $k^d \ll N$, our algorithm can be performed much more efficiently. 

\begin{algorithm}[t]
	\caption{Efficient Mean Shift Clustering.\label{alg:mean_shift}}
	\small
	\begin{algorithmic}[1] %
		\State \textbf{Input:} pixel embeddings $\{x_i\}_{i=1}^N$, hyper-parameters $k$, $d$, $b$, and $T$
		\State initialize $k^d$ anchors uniformly in the embedding space
		\For{$t$ = 1 to $T$}
		\State compute pairwise potential term $p_{ij}^t$ with Eq.~(\ref{eq:pairwise_potential})
		\State conduct mean shift for each anchor with Eq.~(\ref{eq:mean_shift})
		\EndFor
		\State merge nearby anchors to form clusters $\tilde{C}$
		\State \textbf{Output:} instance segmentation map $S$ with Eq.~(\ref{eq:soft_assignment})
	\end{algorithmic}
\end{algorithm}

\subsection{Plane Parameter Estimation}
Given an image, the previous stage provides us a plane instance segmentation map. Then we need to infer the 3D parameter for each plane instance. To this end, we further design a plane parameter branch to predict the plane parameter for each pixel. Then, using the instance segmentation map, we aggregate the output of this branch to form an instance-level parameter for each plane instance. 

Specifically, the branch output a $H \times W \times 3$ plane parameter map. Following~\cite{Yang2018Recovering}, we define the plane parameter as $n \in \mathbb{R}^3$. For 3D points $Q$ lies on this plane, we have $n^T Q = 1$.\footnote{We represent a 3D plane by $n \doteq \tilde{n} / d$, where $\tilde{n} \in \mathcal{S}^2$ and $d$ denote the surface normal and plane distance to the origin.} We use L1 loss to supervise the learning of per-pixel plane parameters:
\begin{equation}
	L_{PP} = \frac{1}{N} \sum_{i=1}^N \| n_{i} - n^*_{i} \|,
\end{equation}
where $n_i$ is the predicted plane parameter and $n^*_{i}$ is the ground truth plane parameter for $i$-th pixel. 

\PAR{Instance-aware pooling.}
In practice, we find that pixel-level parameter supervision is not sufficient, as it may not produce consistent outputs across the entire plane instance. Therefore we propose to further aggregate the pixel-level parameters into an instance-level parameter: 
\begin{equation}
	n_j = \frac{1}{Z_j} \sum_{i=1}^N S_{ij} \cdot n_i,
\end{equation}
where $Z_j = \sum_{i=1}^N S_{ij}$ is a normalization constant. It acts like a global average pooling but with different attention for different plane instances.

Following~\cite{Yang2018Recovering}, we enforce the instance-level parameter to be consistent with the scene geometry. To be specific, we compare the depth map inferred from the plane parameter with the ground truth depth map using the following loss:
\begin{equation}
	L_{IP} = \frac{1}{N \tilde{C}} \sum_{j=1}^{\tilde{C}} \sum_{i=1}^N S_{ij}\cdot \| n_j^T Q_{i} - 1 \|,
\end{equation}
where $Q_{i}$ is the 3D point at pixel $i$ inferred from ground truth depth map.

Note that our approach to plane parameter estimation is different from previous methods~\cite{Liu2018PlaneNet,Yang2018Recovering}. Those methods first predict plane parameter and then associate each pixel with a particular plane parameter. In contrast, we first group pixels into plane instances and then estimate the parameter for each plane instance. We argue that our approach is more adequate because segmentation can uniquely determine an instance. 

Finally, to simultaneously infer plane instance segmentation and plane parameters, the overall training loss of our method is:
\begin{equation}
	L = L_S + L_E + L_{PP} +  L_{IP}.
\end{equation}

\section{Experiments}

In this section, we conduct experiments to evaluate the performance of the proposed method on two public datasets: ScanNet~\cite{Dai2017ScanNet} and NYUv2~\cite{Silberman2012Indoor}. Due to space limitations, we refer readers to appendix for additional experiment results, including ablation studies about the mean shift clustering algorithm and plane parameter estimation.

\subsection{Implementation Details}
We implement our model with PyTorch~\cite{Paszke2017PyTorch}. We use Adam optimizer~\cite{Kingma2015Adam} with a learning rate of $10^{-4}$ and a weight decay of $10^{-5}$. The batch size is set to 16. The network is trained for 50 epochs on one NVIDIA TITAN XP GPU device. We train the network with margins $\delta_{\textrm{v}}=0.5$, $\delta_{\textrm{d}}=1.5$. We set the embedding dimension $d=2$, number of anchors per dimension $k=10$, and the bandwidth $b=\delta_{\textrm{v}}$ in the mean shift clustering algorithm. The number of iterations $T$ is set to $5$ in training and set to $10$ in testing. Our model is trained in an end-to-end manner.

\subsection{Results on ScanNet Dataset}
We first evaluate our method on ScanNet dataset~\cite{Dai2017ScanNet} generated by~\cite{Liu2018PlaneNet}. The ground truth is obtained by fitting planes to a consolidated mesh of ScanNet and project them back to individual frames. The generating process also incorporates semantic annotations from ScanNet. The resulting dataset contains 50,000 training and 760 testing images with resolution $256 \times 192$. 

\begin{figure*}[t]
	\centering
	\setlength{\tabcolsep}{0.1em}
	\renewcommand{\arraystretch}{0.1}
	\hfill{}\hspace*{-0.5em}
	\begin{tabular}{cc|cc}
		\includegraphics[width=0.25\textwidth]{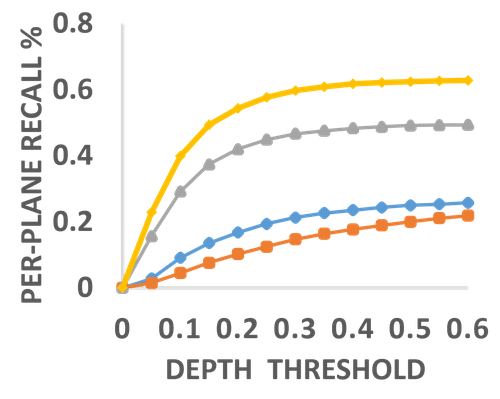}
		&\includegraphics[width=0.25\textwidth]{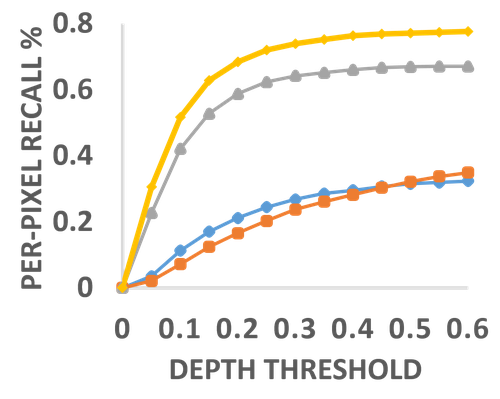}
		&\includegraphics[width=0.25\textwidth]{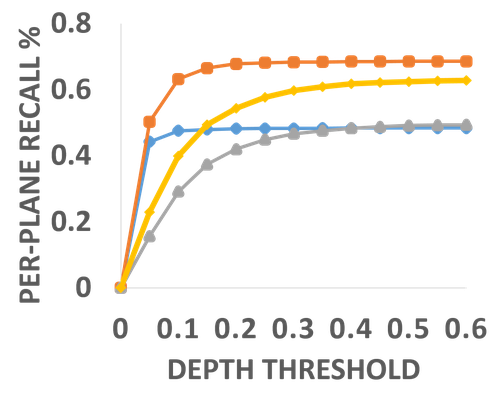}
		&\includegraphics[width=0.25\textwidth]{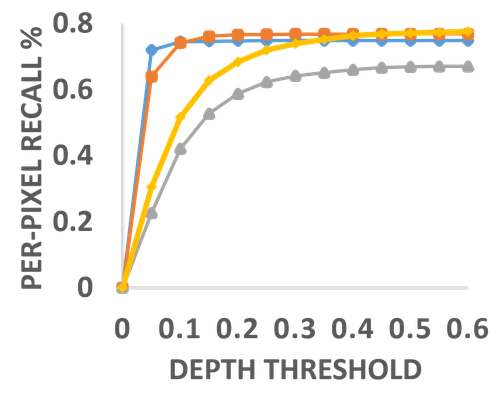}
		\tabularnewline
		\multicolumn{2}{c|}{\includegraphics[width=0.5\textwidth]{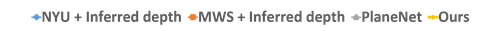}}
		&\multicolumn{2}{c}{\includegraphics[width=0.5\textwidth]{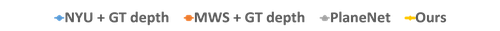}}
	\end{tabular}\hfill{}
	\vspace{-3mm}
	\caption{Plane and pixel recalls on the ScanNet dataset. Please see the appendix for exact numbers.\label{fig:scannet:recall}}
	\vspace{-2mm}
\end{figure*}

\begin{figure*}[t]
	\centering
	\setlength{\tabcolsep}{0.1em}
	\renewcommand{\arraystretch}{0.1}
	\hfill{}\hspace*{-0.5em}
	\begin{tabular}{c|cccccc}
		\begin{turn}{90}
			{\footnotesize{}\hspace{1.5em}}\textcolor{black}{\footnotesize{}Input image}
		\end{turn}
		&\includegraphics[width=0.16\textwidth]{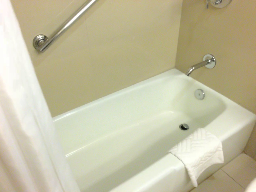}
		&\includegraphics[width=0.16\textwidth]{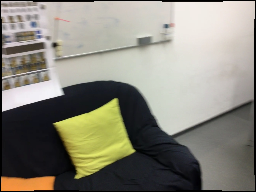} 
		&\includegraphics[width=0.16\textwidth]{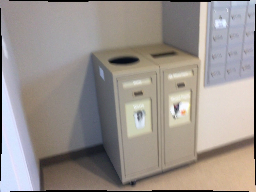} 
		&\includegraphics[width=0.16\textwidth]{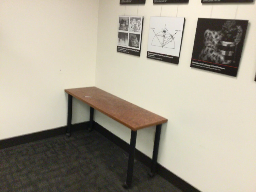}
		&\includegraphics[width=0.16\textwidth]{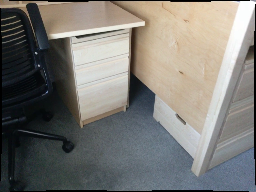}
		&\includegraphics[width=0.16\textwidth]{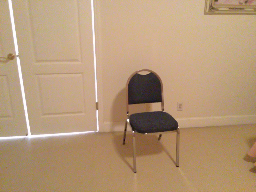}
		\tabularnewline
		\multicolumn{1}{c}{} & & & &
		\tabularnewline
		\begin{turn}{90}
			{\footnotesize{}\hspace{1.2em}}\textcolor{black}{\footnotesize{}Segmentation}
		\end{turn}
		&\includegraphics[width=0.16\textwidth]{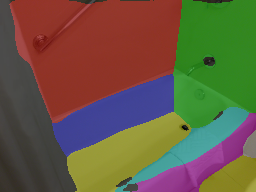}
		&\includegraphics[width=0.16\textwidth]{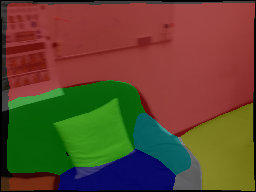} 
		&\includegraphics[width=0.16\textwidth]{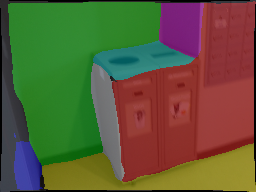} 
		&\includegraphics[width=0.16\textwidth]{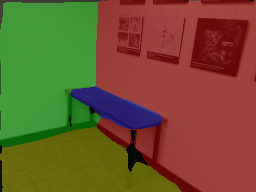}
		&\includegraphics[width=0.16\textwidth]{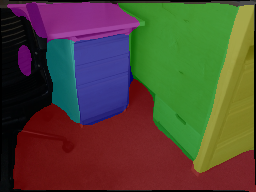}
		&\includegraphics[width=0.16\textwidth]{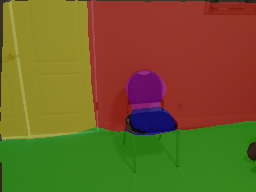}
		\tabularnewline
		\multicolumn{1}{c}{} & & & &
		\tabularnewline
		\begin{turn}{90}
			{\footnotesize{}\hspace{1.5em}}\textcolor{black}{\footnotesize{}Depth map}
		\end{turn}
		&\includegraphics[width=0.16\textwidth]{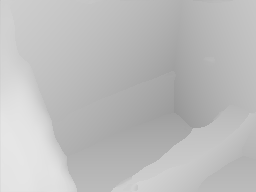}
		&\includegraphics[width=0.16\textwidth]{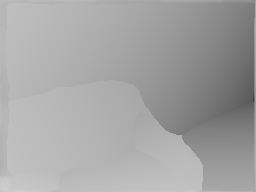} 
		&\includegraphics[width=0.16\textwidth]{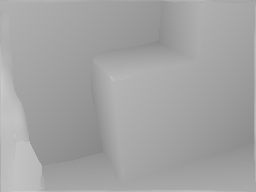}
		&\includegraphics[width=0.16\textwidth]{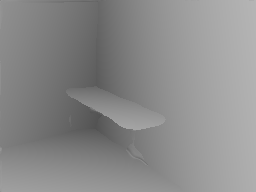}
		&\includegraphics[width=0.16\textwidth]{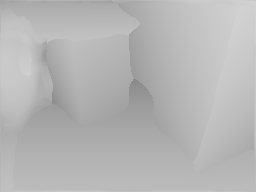}
		&\includegraphics[width=0.16\textwidth]{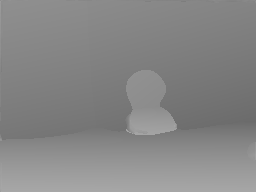}
		\tabularnewline
		\multicolumn{1}{c}{} & & & &
		\tabularnewline
		\begin{turn}{90}
			{\footnotesize{}\hspace{0.2em}}\textcolor{black}{\footnotesize{}Planar 3D model}
		\end{turn}
		&\includegraphics[width=0.16\textwidth]{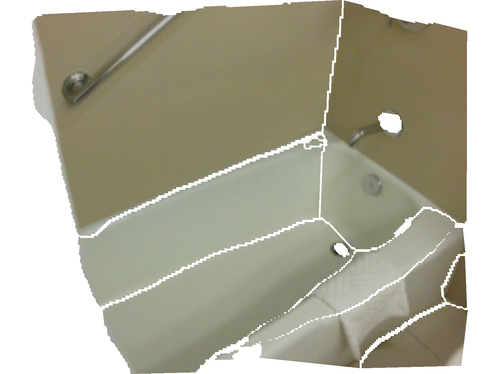}
		&\includegraphics[width=0.16\textwidth]{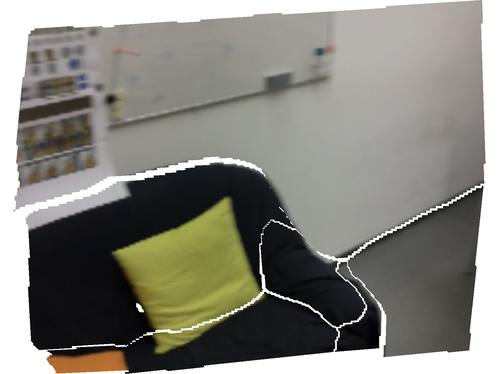}
		&\includegraphics[width=0.16\textwidth]{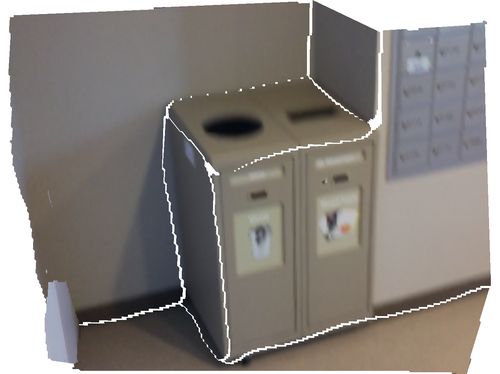}
		&\includegraphics[width=0.16\textwidth]{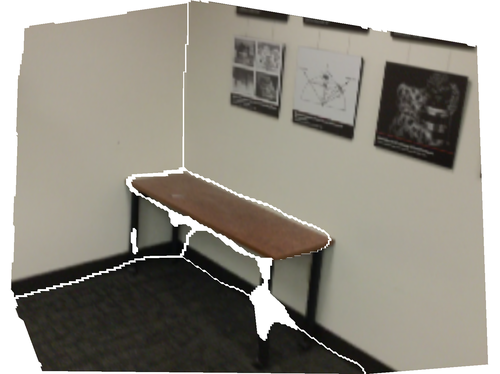}	
		&\includegraphics[width=0.16\textwidth]{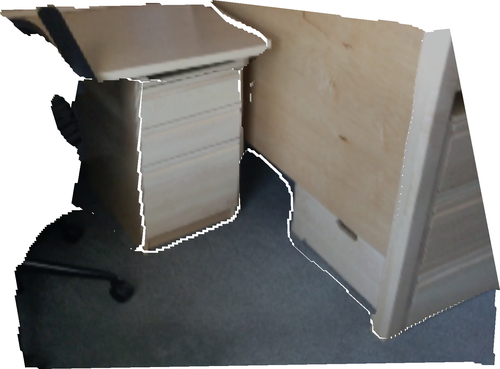}
		&\includegraphics[width=0.16\textwidth]{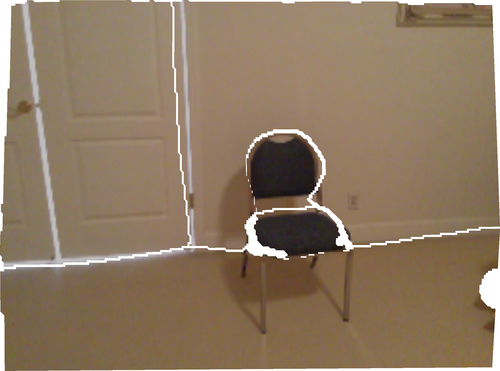}
	\end{tabular}\hfill{}
	\vspace{-3mm}
	\caption{Piece-wise planar 3D reconstruction results on the ScanNet dataset. In the plane instance segmentation results, black color indicates non-planar regions.\label{fig:scannet:visual}}
	\vspace{-5mm}
\end{figure*}`

\PAR{Methods for comparison.}
We compare our method with the recent CNN-based method PlaneNet~\cite{Liu2018PlaneNet}, and two bottom-up methods NYU-Toolbox~\cite{Silberman2012Indoor} and Manhattan World Stereo (MWS)~\cite{Furukawa2009Manhattan}.\footnote{We obtain the implementation of these methods from PlaneNet~\cite{Liu2018PlaneNet} at \url{https://github.com/art-programmer/PlaneNet}.} NYU-Toolbox~\cite{Silberman2012Indoor} is a popular plane detection algorithm that uses RANSAC to extracts plane hypotheses and Markov Random Field (MRF) to optimize plane segmentation. Manhattan World Stereo (MWS)~\cite{Furukawa2009Manhattan} employs Manhattan world assumption for plane extraction and utilizes vanishing lines in the pairwise terms of MRF.  For bottom-up methods, we use the same network architecture as ours to predict pixel-level depth map. Following~\cite{Laina2016Deeper}, we minimize the berHu loss during training. Alternatively, we also use ground truth depth map as input for these methods.

\PAR{Evaluation metric.}
Following~\cite{Liu2018PlaneNet}, we use plane and pixel recalls as our evaluation metrics. The plane recall is the percentage of correctly predicted ground truth planes, and the pixel recall is the percentage of pixels within the correctly predicted planes. A ground-truth plane is considered correctly predicted if i) one of the predicted planes has more than 0.5 intersection-over-union (IOU) score, and ii) the mean depth difference over the overlapping region is less than a threshold, which varies from 0.05m to 0.6m with an increment of 0.05m. In addition, we also use surface normal difference as the threshold in our experiment. 

\begin{figure*}[t]
	\centering
	\setlength{\tabcolsep}{0.1em}
	\renewcommand{\arraystretch}{0.5}
	\hfill{}\hspace*{-0.5em}
	\begin{tabular}{cccccc}
		\includegraphics[width=0.16\textwidth]{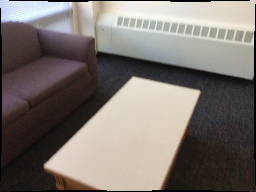}
		&\includegraphics[width=0.16\textwidth]{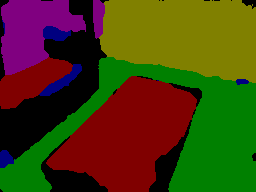}
		&\includegraphics[width=0.16\textwidth]{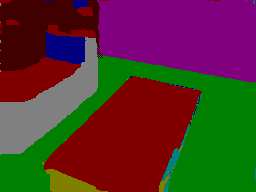}
		&\includegraphics[width=0.16\textwidth]{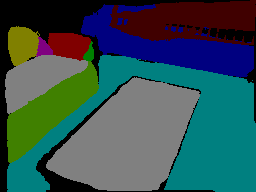}
		&\includegraphics[width=0.16\textwidth]{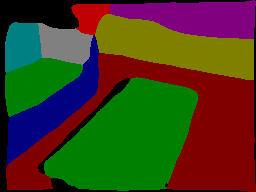}
		&\includegraphics[width=0.16\textwidth]{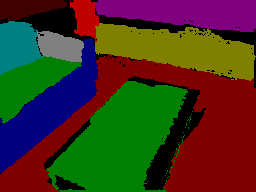}
		\tabularnewline
		\includegraphics[width=0.16\textwidth]{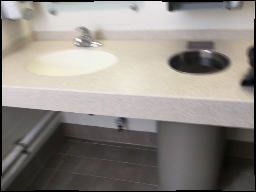}
		&\includegraphics[width=0.16\textwidth]{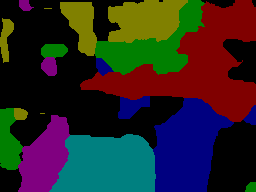}
		&\includegraphics[width=0.16\textwidth]{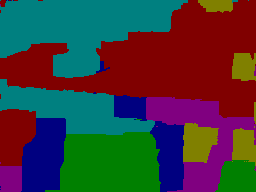}
		&\includegraphics[width=0.16\textwidth]{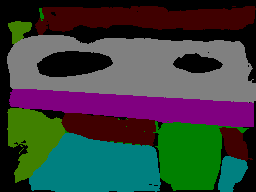}
		&\includegraphics[width=0.16\textwidth]{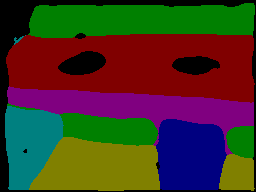}
		&\includegraphics[width=0.16\textwidth]{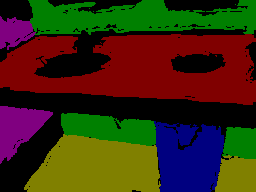}
		\tabularnewline
		\includegraphics[width=0.16\textwidth]{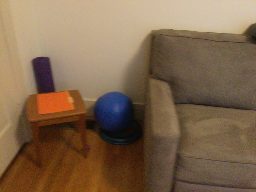}
		&\includegraphics[width=0.16\textwidth]{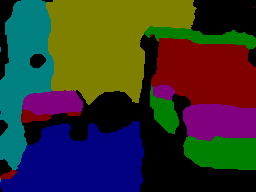}
		&\includegraphics[width=0.16\textwidth]{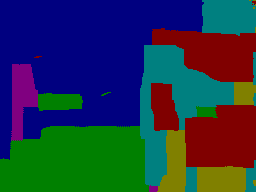}
		&\includegraphics[width=0.16\textwidth]{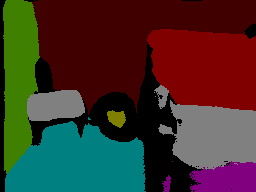}
		&\includegraphics[width=0.16\textwidth]{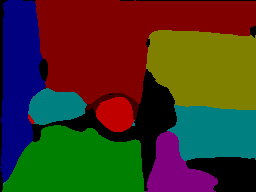}
		&\includegraphics[width=0.16\textwidth]{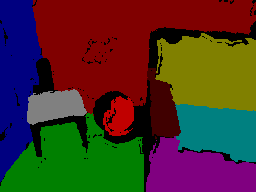}
		\tabularnewline
		\includegraphics[width=0.16\textwidth]{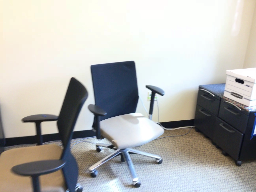}
		&\includegraphics[width=0.16\textwidth]{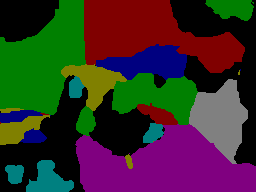}
		&\includegraphics[width=0.16\textwidth]{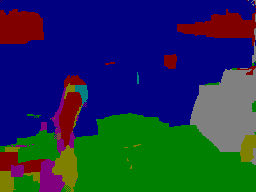}
		&\includegraphics[width=0.16\textwidth]{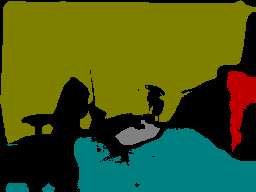}
		&\includegraphics[width=0.16\textwidth]{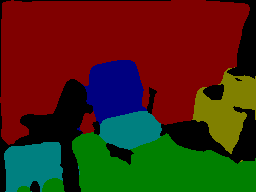}
		&\includegraphics[width=0.16\textwidth]{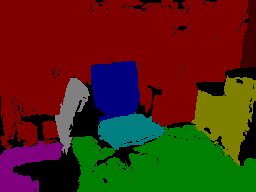}
		\tabularnewline
		\includegraphics[width=0.16\textwidth]{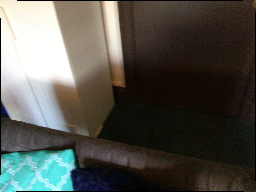}
		&\includegraphics[width=0.16\textwidth]{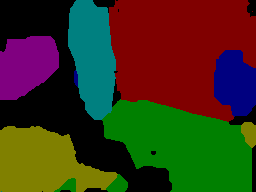}
		&\includegraphics[width=0.16\textwidth]{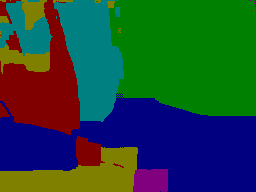}
		&\includegraphics[width=0.16\textwidth]{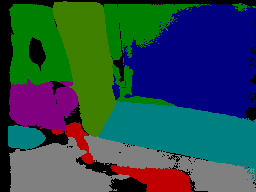}
		&\includegraphics[width=0.16\textwidth]{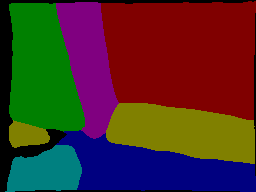}
		&\includegraphics[width=0.16\textwidth]{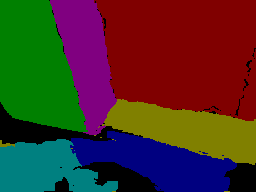}
		\tabularnewline
		{\small{}Input image} 
		&{\small{}NYU-Toolbox~\cite{Silberman2012Indoor}} &{\small{}MWS~\cite{Furukawa2009Manhattan}}
		&{\small{}PlaneNet~\cite{Liu2018PlaneNet}} 
		&{\small{}Ours} 
		&{\small{}Ground truth}
	\end{tabular}\hfill{}
	\vspace{-2mm}
	\caption{Plane instance segmentation results on the ScanNet dataset.\label{fig:scannet:compare}}
	\vspace{-4mm}
\end{figure*}

\PAR{Quantitative evaluation.}
Figure~\ref{fig:scannet:recall} shows the pixel and plane recalls of all methods. As shown in the first two plots, our method significantly outperforms all competing methods when inferred depth maps are used. Furthermore, we achieve competitive or better results even when the bottom-up methods are provided with the ground truth depth maps, as shown in the last two plots. This clearly demonstrates the effectiveness of our method. Furthermore, we obtain consistent results when the surface normal difference is adopted as the threshold (see appendix). 

\begin{table}[t]
	\centering
	\caption{Runtime comparison ($^*$ denotes CPU time).\label{tab:runtime}}
	\vspace{-2mm}
	\begin{tabular}{c|cccc}
		\multirow{2}{*}{\small{}Method} &{\small{}NYU-Toolbox} &{\small{}MWS} &{\small{}PlaneNet} &\multirow{2}{*}{\small{}Ours}
		\tabularnewline
		&{\small{}\cite{Silberman2012Indoor}} &{\small{}\cite{Furukawa2009Manhattan}} &{\small{}\cite{Liu2018PlaneNet}}
		\tabularnewline
		\hline \hline
		{\small{}FPS} &{\small{}0.14$^*$} &{\small{}0.05$^*$} &{\small{}1.35} &\textbf{\small{}32.26}
	\end{tabular}
	\vspace{-5mm}
\end{table}

\PAR{Qualitative evaluation.}
Figure~\ref{fig:scannet:visual} shows our reconstruction results for a variety of scenes. The qualitative comparisons against existing methods on plane instance segmentation are shown in Figure~\ref{fig:scannet:compare}. We make the following observations: i) All methods perform well in simple scenarios (\eg, the first row). ii) PlaneNet~\cite{Liu2018PlaneNet} and our method produce significantly better segmentation results in most cases (\eg, the second and the third row). The poor performance by bottom-up methods is likely due to the noise in the predicted depth maps. In such cases, it is hard to select a proper threshold to distinguish inliers (\ie, points on a plane) and outliers during the grouping stage. iii) PlaneNet sometimes misses small planes (\eg, the chairs in the fourth row) or incorrectly merges multiple planes (\eg, the cabinet and the door in the fifth row), while our approach is more robust in those cases. This is probably due to the assumption of a fixed number of planes in PlaneNet. Our approach is not restricted to such an assumption, thus performs better in detecting structures at different scales. 

\PAR{Speed.} 
Table~\ref{tab:runtime} shows the runtime comparison results with other methods on the ScanNet dataset. All timings are measured on the same computing platform with Xeon E5-2630 @2.2GHz (20 cores) and a single NVIDIA TITAN XP GPU. Our method achieves the fastest speed of 32.26 fps on a single GPU, making it suitable for many real-time applications such as visual SLAM.

\begin{table}[t]
	\centering
	\caption{Plane instance segmentation results on the NYUv2 test set.\label{tab:nyuv2}}
	\vspace{-4mm}
	\begin{tabular}{c|ccc}
		{\small{}Method} &{\small{}RI $\uparrow$} &{\small{}VI $\downarrow$} &{\small{}SC $\uparrow$}
		\tabularnewline
		\hline \hline
		{\small{}GT Depth + NYU-Toolbox~\cite{Silberman2012Indoor}} &{\small{}0.875} &{\small{}1.284} &{\small{}0.544}
		\tabularnewline
		{\small{}PlaneNet~\cite{Liu2018PlaneNet}} &{\small{}0.723} &{\small{}1.932} &{\small{}0.404}
		\tabularnewline
		{\small{}Ours} &\textbf{\small{}0.888} &\textbf{\small{}1.380} &\textbf{\small{}0.519}
	\end{tabular}
	\vspace{-4mm}
\end{table}

\begin{table*}[t]
	\centering
	\caption{Comparison of depth prediction accuracy on the NYUv2 test set.\label{tab:nyuv2:depth}}
	\vspace{-4mm}
	\begin{tabular}{c|ccccc|ccc}
		\multirow{2}{*}{\small{}Method} &\multicolumn{5}{c|}{\small{}Lower the better} &\multicolumn{3}{c}{\small{}Higher the better} 
		\tabularnewline
		&{\small{}Rel} &{\small{}Rel(sqr)} &{\small{}log$_{10}$} &{\small{}RMSE$_{iin}$} &{\small{}RMSE$_{\log}$} &{\small{}1.25} &{\small{}1.25$^2$} &{\small{}1.25$^3$}
		\tabularnewline
		\hline \hline
		{\small{}Eigen-VGG~\cite{Eigen2015Predicting}} &{\small{}0.158} &{\small{}0.121} &{\small{}0.067} &{\small{}0.639} &{\small{}0.215} &{\small{}77.1} &{\small{}95.0} &{\small{}98.8}
		\tabularnewline
		{\small{}SURGE~\cite{Wang2016SURGE}} &{\small{}0.156} &{\small{}0.118} &{\small{}0.067} &{\small{}0.643} &{\small{}0.214} &{\small{}76.8} &{\small{}95.1} &{\small{}98.9}
		\tabularnewline
		{\small{}FCRN~\cite{Laina2016Deeper}} &{\small{}0.152} &{\small{}0.119} &{\small{}0.072} &{\small{}0.581} &{\small{}0.207} &{\small{}75.6} &{\small{}93.9} &{\small{}98.4}
		\tabularnewline
		{\small{}PlaneNet~\cite{Liu2018PlaneNet}} &{\small{}0.142} &{\small{}0.107} &{\small{}0.060} &{\small{}0.514} &{\small{}0.179} &{\small{}81.2} &{\small{}95.7} &{\small{}98.9}
		\tabularnewline
		{\small{}Ours (depth-direct)} &\textbf{\small{}0.134} &\textbf{\small{}0.099} &\textbf{\small{}0.057} &\textbf{\small{}0.503} &\textbf{\small{}0.172} &\textbf{\small{}82.7} &\textbf{\small{}96.3} &\textbf{\small{}99.0}
		\tabularnewline
		{\small{}Ours} &{\small{}0.141} &{\small{}0.107} &{\small{}0.061} &{\small{}0.529} &{\small{}0.184} &{\small{}81.0} &{\small{}95.7} &\textbf{\small{}99.0}
	\end{tabular}
	\vspace{-5mm}
\end{table*}

\subsection{Results on NYUv2 Dataset}
We further evaluate the performance of our method on the NYUv2 dataset~\cite{Silberman2012Indoor}, which contains 795 training images and 654 test images. Specifically, we conduct experiments to examine i) the generalizability of our learnt embedding on plane instance segmentation, and ii) the depth prediction accuracy of our method.

\PAR{Plane instance segmentation.} 
In this experiment, we directly use PlaneNet and our model trained on the ScanNet dataset to predict plane instances on the NYUv2 dataset. Following~\cite{Liu2018PlaneNet}, we generate ground truth plane instances in the test images by first fitting plane in each semantic instance using RANSAC and further merging two planes if the mean distance is below 10cm. For quantitative evaluation, we employ three popular metrics in segmentation~\cite{ArbeiAez2011Contour, Yang2018Recovering}: Rand index (RI), variation of information (VI), and segmentation covering (SC). As shown in Table~\ref{tab:nyuv2}, our method significantly outperforms PlaneNet in terms of all metrics. This suggests that our embedding-based approach is more generalizable than existing CNN-based method. And our method remains competitive against traditional bottom-up method NYU-Toolbox, even when the latter is provided with the ground truth depth maps. We refer readers to appendix for qualitative results on the NYUv2 dataset. 

\PAR{Depth prediction.}
While our method demonstrates superior performance in piece-wise planar 3D reconstruction, it's also interesting to evaluate the network's capacity for per-pixel depth prediction. For this experiment, we fine-tune our network using the ground truth plane instances we generated on NYUv2 dataset. Table~\ref{tab:nyuv2:depth} compares the accuracy of the depth maps derived from our network output (i.e., the piece-wise planar 3D models) against those generated by standard depth prediction methods. As one can see, our method outperforms or is comparable to all other methods, which further verifies the quality of the 3D planes recovered by our method. 

We have also trained a variant of our network, denoted as ``Ours (depth-direct)'', by fine-tuning the plane parameter prediction branch only using pixel-level supervision. Then, we use this branch to directly predict depth maps. As shown in Table~\ref{tab:nyuv2:depth}, compared to this variant, using our piece-wise planar representation results in slight decrease in depth prediction accuracy, as such a representation sometimes ignores details and small variations in the scene structure.

\subsection{Failure Cases}
\begin{figure}[t]
	\centering
	\setlength{\tabcolsep}{0.1em}
	\renewcommand{\arraystretch}{0.5}
	\hfill{}\hspace*{-0.5em}
	\begin{tabular}{ccc}
		\includegraphics[width=0.33\columnwidth]{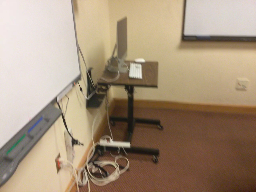}
		&\includegraphics[width=0.33\columnwidth]{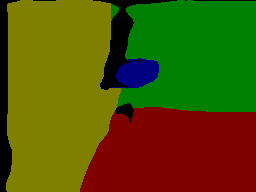}
		&\includegraphics[width=0.33\columnwidth]{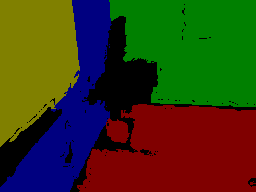}
		\tabularnewline
		\includegraphics[width=0.33\columnwidth]{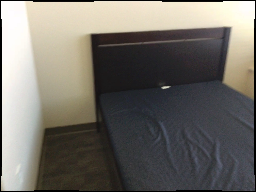}
		&\includegraphics[width=0.33\columnwidth]{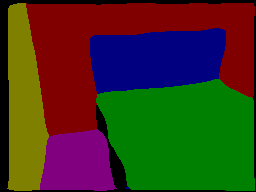}
		&\includegraphics[width=0.33\columnwidth]{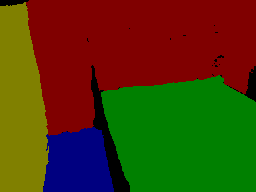}
		\tabularnewline
		\includegraphics[width=0.33\columnwidth]{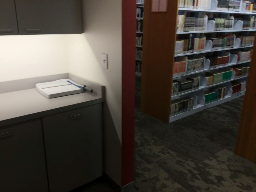}
		&\includegraphics[width=0.33\columnwidth]{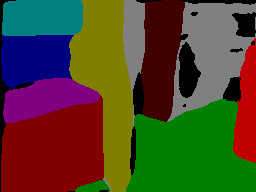}
		&\includegraphics[width=0.33\columnwidth]{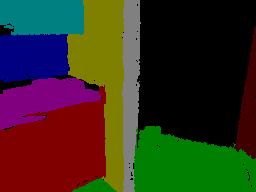}
		\tabularnewline
		{\small{}Input image} 
		&{\small{}Ours} 
		&{\small{}Ground truth}
	\end{tabular}\hfill{}
	\vspace{-2mm}
	\caption{Failure cases.\label{fig:failure}}
	\vspace{-4mm}
\end{figure}

We show some failure cases in Figure~\ref{fig:failure}. In the first example, our method merges the whiteboard and the wall into one plane. This may be because the appearance of these two planes are similar. A possible solution is to separate them by incorporating semantic information. In the second example, our method separates one plane into two planes (wall and headboard) because of the distinct appearance. One can easily merge these two planes using plane parameters in a post-processing step. In the third example, our method fails to segment the whole bookshelf. A possible reason is that the plane instance annotations obtained by fitting are not consistent, \ie, the bookshelf, in this case, is not labeled in the ground truth.

\section{Conclusion}
This paper proposes a novel two-stage method to single-image piece-wise planar 3D reconstruction. Specifically, we learn a deep embedding model to directly segment plane instances in an image, then estimate 3D parameter for each plane by considering pixel-level and instance-level geometric consistencies. The proposed method significantly outperforms the state-of-the-art methods while achieving real-time performance. While the proposed method has demonstrated promising results, it is still far from perfect. Possible future directions include i) incorporating semantic information to improve the reconstruction results, and ii) learning to reconstruct piece-wise planar 3D models from videos by leveraging the theory of multi-view geometry.

\section*{Acknowledgements}
This work was supported by NSFC \#61502304. Zihan Zhou was supported by NSF award \#1815491. 

{\small
\bibliographystyle{ieee}
\bibliography{library}
}

\appendix

\begin{appendices}

In the appendix, we first present the details of our network architecture. We then show some ablation studies of our proposed method. Finally we report additional quantitative and qualitative results on two public datasets: ScanNet~\cite{Dai2017ScanNet} and NYUv2~\cite{Silberman2012Indoor}. 

\section{Architecture}
Our encoder is an extended version of ResNet-101-FPN~\cite{Lin2017Feature}. We add two lateral connections and top-down pathways to the original FPN, and the size of resulting feature map from the encoder is $64 \times 192 \times 256$. Three decoders, \ie, plane segmentation decoder, plane embedding decoder, and plane parameter decoder in Figure~\ref{fig:pipeline}, all share this feature map. Each decoder simply contains a $1 \times 1$ convolutional layer. The architecture is shown in Table~\ref{tab:arch}. 

\begin{table}[h]
	\centering
	\caption{Network architecture.\label{tab:arch}}
	\vspace{-2mm}
	\begin{tabular}{c|c|c}
		{\small{}Stage} &{\small{}Type} &{\small{}Output Size}
		\tabularnewline \hline \hline
		{\small{}Input} & &{\small{}$3 \times 192 \times 256$}
		\tabularnewline
		\multirow{2}{*}{\small{}Encoder} &
		\multirow{2}{*}{
			\begin{tabular}{c}
				{\scriptsize{}Extended} \tabularnewline
				{\scriptsize{}ResNet-101-FPN}
		\end{tabular}}
		&\multirow{2}{*}{{\small{}$64 \times 192 \times 256$}}
		\tabularnewline
		& & 
		\tabularnewline \hline
		{\small{}Plane segm. decoder} &{\small{}$1\times1$ Conv} &{\small{}$1 \times 192 \times 256$}
		\tabularnewline
		{\small{}Plane embed. decoder} &{\small{}$1\times1$ Conv} &{\small{}$2 \times 192 \times 256$}
		\tabularnewline
		{\small{}Plane param. decoder} &{\small{}$1\times1$ Conv} &{\small{}$3 \times 192 \times 256$}
		\tabularnewline
	\end{tabular}
\end{table}

\section{Ablation Studies}
In this section, we run a number of ablation studies to validate our method. We use plane recall and pixel recall at 0.05m and 0.6m to evaluate the performance of our methods on the ScanNet test set. 

\begin{figure}[t]
	\centering
	\setlength{\tabcolsep}{0.1em}
	\hfill{}\hspace*{-0.5em}
	\begin{tabular}{cc}
		\includegraphics[width=0.5\columnwidth]{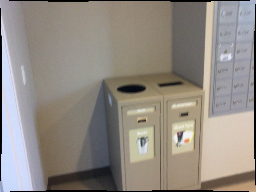} 
		&\includegraphics[width=0.5\columnwidth]{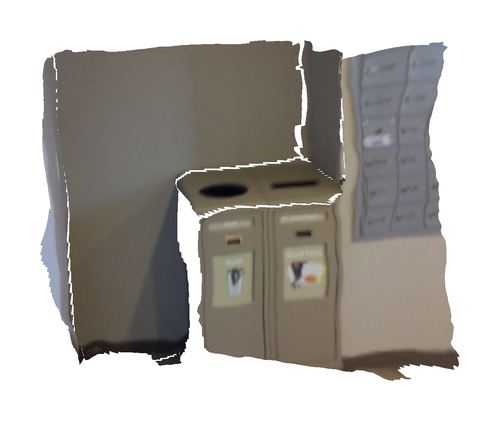}
		\tabularnewline
		{\small{}Input image} 
		&{\small{}Supervision: $L_{PP} + L_{IP}$} 
		\tabularnewline
		\includegraphics[width=0.5\columnwidth]{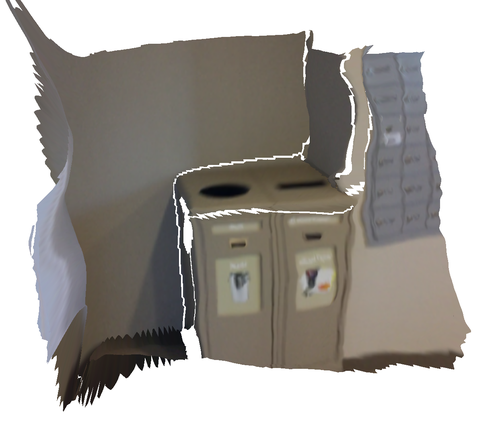} 
		&\includegraphics[width=0.5\columnwidth]{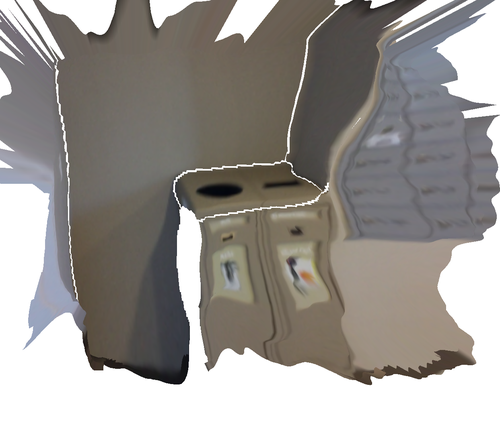}
		\tabularnewline
		{\small{}Supervision: $L_{PP}$} 
		&{\small{}Supervision: $L_{IP}$} 
	\end{tabular}\hfill{}
	\caption{Visualization about plane parameter supervision. Note that all results are reconstructed with the depth maps inferred from pixel-level plane parameters. The results with both supervisions are more consistent and stable.\label{fig:ablation:param}}
\end{figure}

\PAR{Plane parameter.}
To evaluate the effectiveness of our plane parameter supervisions, we remove either pixel-level parameter supervision $L_{PP}$ or instance-level parameter supervision $L_{IP}$ in this experiment. As shown in Table~\ref{tab:ablation:param}, both terms play an important role in estimating the scene geometry. Figure~\ref{fig:ablation:param} further visualizes the reconstruction results derived from the predicted pixel-level parameters. We make the following observations: i) the network with pixel-level parameter supervision $L_{PP}$ only produces inconsistent parameters across the entire plane; ii) the network with instance-level parameter supervision $L_{IP}$ only generates reasonably good results w.r.t. the whole scene geometry, but fails produce accurate predictions at pixel level (\eg, the boundary of each plane); iii) with both supervisions, the results are more consistent and stable. 

\begin{table}[t]
	\centering
	\caption{Ablation study of plane parameter supervisions on the ScanNet test set. The $\checkmark$ indicates the enabled supervision.\label{tab:ablation:param}}
	\vspace{-2mm}
	\begin{tabular}{cc|cc|cc}
		\multicolumn{2}{c|}{\small{}Supervision} 
		&\multicolumn{2}{c|}{\small{}Per-plane recall} &\multicolumn{2}{|c}{\small{}Per-pixel recall}
		\tabularnewline
		{\small{}$L_{PP}$} &{\small{}$L_{IP}$} &{\small{}@0.05} &{\small{}@0.60} &{\small{}@0.05} &{\small{}@0.60}
		\tabularnewline
		\hline \hline
		\checkmark & &{\small{}20.18} &{\small{}61.16} &{\small{}24.82} &{\small{}75.10} 
		\tabularnewline
		&\checkmark &{\small{}10.78} &{\small{}62.04} &{\small{}15.72} &{\small{}76.61} 
		\tabularnewline
		\checkmark &\checkmark &\textbf{\small{}22.93} &\textbf{\small{}62.93} &\textbf{\small{}30.59} &\textbf{\small{}77.86}
		\tabularnewline
	\end{tabular}
\end{table}

\PAR{Clustering.}
To validate the efficiency of our mean shift clustering algorithm, we compare our algorithm with vanilla mean shift algorithm in \emph{scikit-learn}~\cite{Pedregosa2011Scikitlearn}. We further analyze the effect of two hyper-parameters: i) the number of anchors per dimension $k$, ii) the number of iteration $T$ in testing. Experimental results are shown in Table~\ref{tab:ablation:cluster}. All timings are recorded on the same computing platform with a 2.2GHz 20-core Xeon E5-2630 CPU and a single NVIDIA TITAN Xp GPU. Our proposed method is more efficient, achieving 30 fps on a single GPU. Further, our proposed method is robust to hyper-parameter selection. 

\begin{table*}[t]
	\centering
	\caption{Ablation study of clustering on the ScanNet test set. The $^*$ indicates CPU time (with 20 cores). Our method is more efficient and is robust to hyper-parameters selection.\label{tab:ablation:cluster}}
	\begin{tabular}{c|c|c|cc|cc|c}
		\multirow{2}{*}{Variant} &\multicolumn{2}{|c|}{Hyper-param.}
		&\multicolumn{2}{|c|}{Per-plane recall} &\multicolumn{2}{|c|}{Per-pixel recall}
		&{Speed} 
		\tabularnewline
		&{\small{}k} &{\small{}T} &{\small{}@0.05} &{\small{}@0.60} &{\small{}@0.05} &{\small{}@0.60} &{\small{}(FPS)}
		\tabularnewline
		\hline \hline
		scikit-learn &- &- &{22.85} &{63.13} &{30.18} &{76.09} &{2.86$^*$} 
		\tabularnewline
		\hline
		\multirow{6}{*}{Ours} &10 &10 &{22.96} &{62.89} &{30.64} &{77.70} &{32.26} 
		\tabularnewline
		&20 &10 &{22.97} &{62.96} &{30.62} &\textbf{77.80} &{22.19} 
		\tabularnewline
		&50 &10 &{23.05} &{63.11} &{30.71} &{77.73} &{6.69} 
		\tabularnewline
		&10 &5 &\textbf{23.28} &{63.65} &\textbf{30.77} &{77.70} &\textbf{36.10} 
		\tabularnewline
		&20 &5 &{23.18} &\textbf{63.72} &{30.68} &{77.58} &{24.39} 
		\tabularnewline
		&50 &5 &{22.94} &{63.35} &{30.41} &{76.85} &{8.08} 
		\tabularnewline
	\end{tabular}
\end{table*}

\begin{figure*}[t]
	\centering
	\setlength{\tabcolsep}{0.1em}
	\renewcommand{\arraystretch}{0.1}
	\hfill{}\hspace*{-0.5em}
	\begin{tabular}{cc|cc}
		\includegraphics[width=0.5\columnwidth]{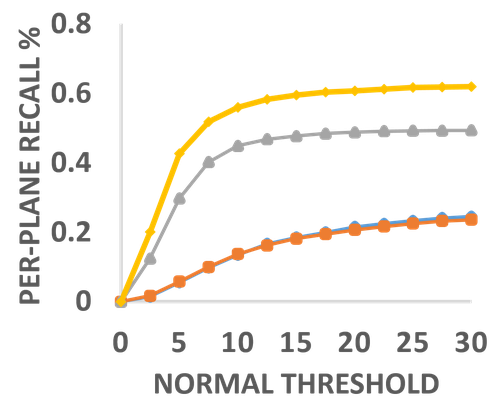}
		&\includegraphics[width=0.5\columnwidth]{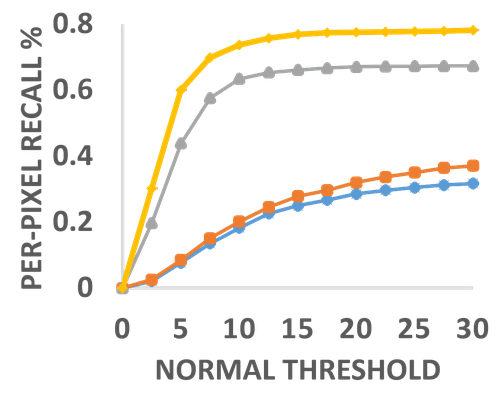}
		&\includegraphics[width=0.5\columnwidth]{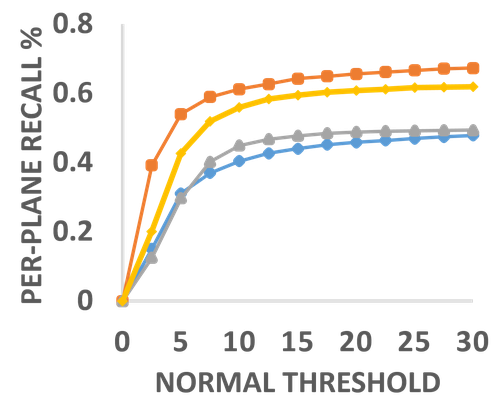}
		&\includegraphics[width=0.5\columnwidth]{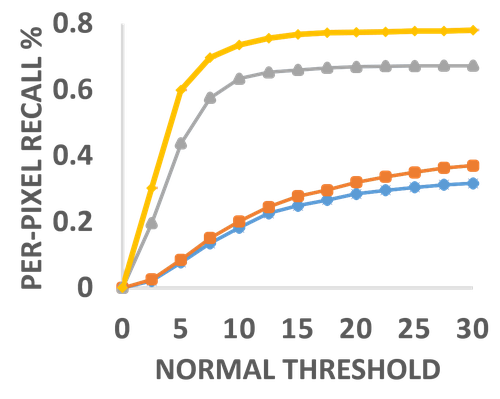}
		\tabularnewline
		\multicolumn{2}{c|}{\includegraphics[width=\columnwidth]{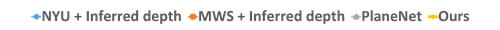}}
		&\multicolumn{2}{c}{\includegraphics[width=\columnwidth]{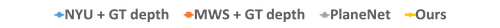}}
	\end{tabular}\hfill{}
	\caption{Plane and pixel recall curves with normal difference as threshold on the ScanNet dataset. Our method obtains consistent results when surface normal difference is adopted as threshold.\label{fig:scannet:normal}}
	\vspace{-2mm}
\end{figure*}

\section{More Results}
In this section, we show more results on the ScanNet and NYUv2 datasets. 

\begin{figure}[t]
\centering
\includegraphics[width=0.9\linewidth]{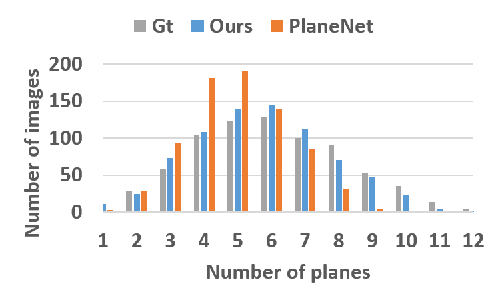}
\vspace{-2mm}
\caption{The number of images versus the number of planes in the image.\label{fig:stat}}
\vspace{-2mm}
\end{figure}

\PAR{Statistics on the number of detected planes.}
We show some statistics on the number of planes in Figure~\ref{fig:stat}. The histogram illustrates the number of images versus the number of planes. We make the following observations: i) Due to the limitation of a fixed number of planes (\ie, 10 planes in PlaneNet), PlaneNet~\cite{Liu2018PlaneNet} cannot detect all the planes if there are more than 10 planes in the image. ii) Our method is more consistent with the ground truth than PlaneNet. 

\PAR{Quantitative evaluation.}
We further provide the experiment of depth prediction without fine-tuning on the NYUv2 dataset in Table~\ref{tab:nyuv2:depth_wo}. The results show our method generalizes well. 

Besides using depth as threshold, we also use surface normal difference (in degrees) between the predicted plane and ground truth plane as threshold. The threshold varies from $0^\circ$ to $30^\circ$ with an increment of $2.5^\circ$. As shown in Figure~\ref{fig:scannet:normal}, the results are consistent with the results when depth is adopted as threshold. We list the exact numbers of each recall curve in Table~\ref{tab:scannet:recall}. 

\begin{figure*}[t]
	\centering
	\setlength{\tabcolsep}{0.1em}
	\renewcommand{\arraystretch}{0.1}
	\hfill{}\hspace*{-0.5em}
	\begin{tabular}{cccc}
		\includegraphics[width=0.25\textwidth]{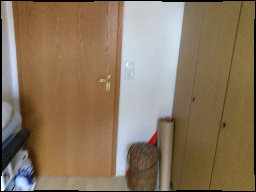}
		&\includegraphics[width=0.25\textwidth]{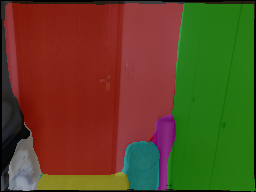}	
		&\includegraphics[width=0.25\textwidth]{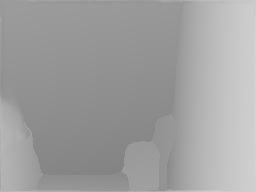}
		&\includegraphics[width=0.25\textwidth]{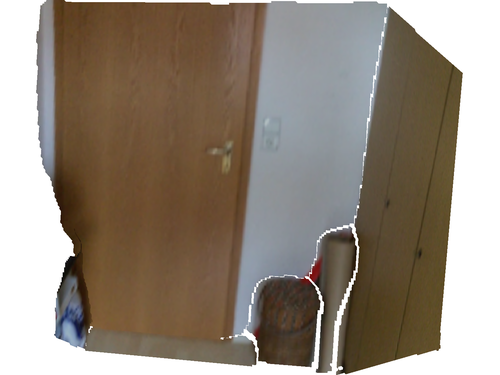}	
		\tabularnewline
		& & & \smallskip{} 
		\tabularnewline
		\includegraphics[width=0.25\textwidth]{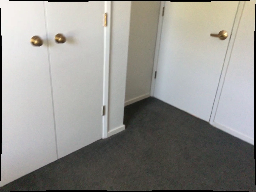}	
		&\includegraphics[width=0.25\textwidth]{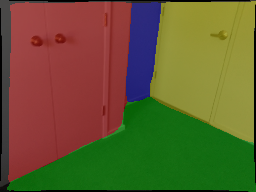}	
		&\includegraphics[width=0.25\textwidth]{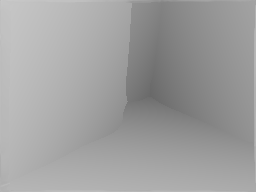}	
		&\includegraphics[width=0.25\textwidth]{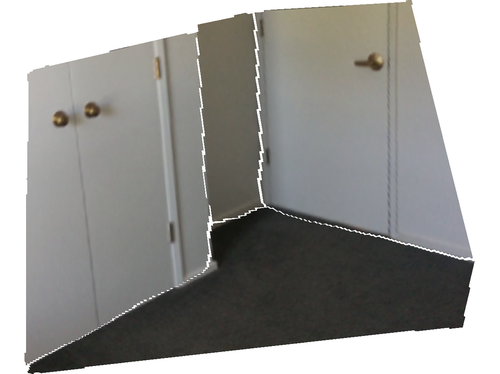}
		\tabularnewline
		& & & \smallskip{} 
		\tabularnewline
		\includegraphics[width=0.25\textwidth]{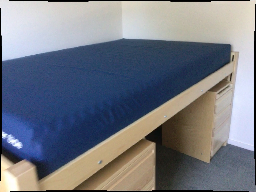}
		&\includegraphics[width=0.25\textwidth]{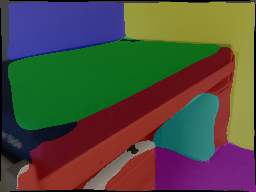}	
		&\includegraphics[width=0.25\textwidth]{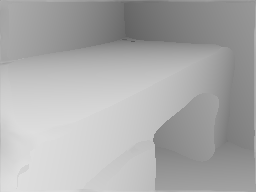}
		&\includegraphics[width=0.25\textwidth]{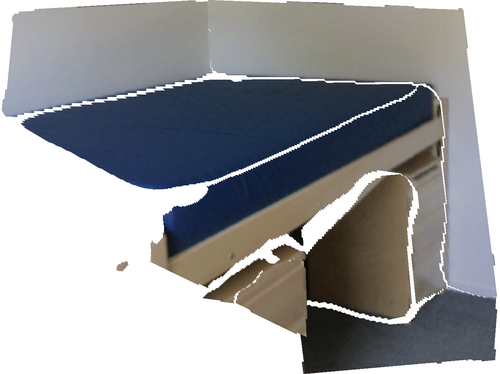}
		\tabularnewline
		& & & \smallskip{} 
		\tabularnewline
		\includegraphics[width=0.25\textwidth]{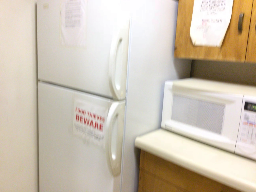}
		&\includegraphics[width=0.25\textwidth]{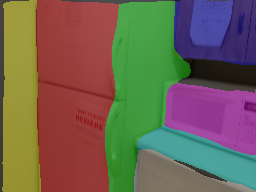}	
		&\includegraphics[width=0.25\textwidth]{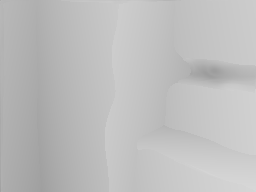}
		&\includegraphics[width=0.25\textwidth]{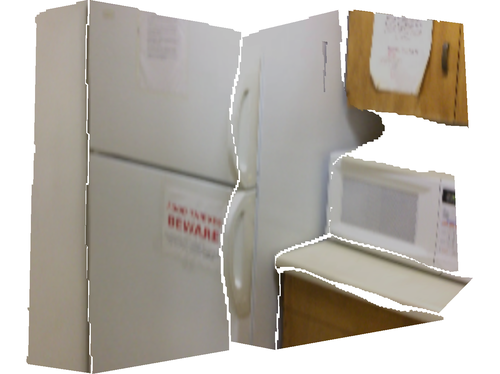}
		\tabularnewline
		& & & \smallskip{} 
		\tabularnewline
		\includegraphics[width=0.25\textwidth]{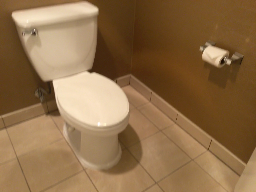}
		&\includegraphics[width=0.25\textwidth]{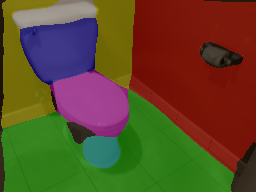}	
		&\includegraphics[width=0.25\textwidth]{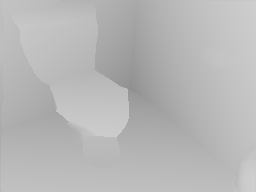}
		&\includegraphics[width=0.25\textwidth]{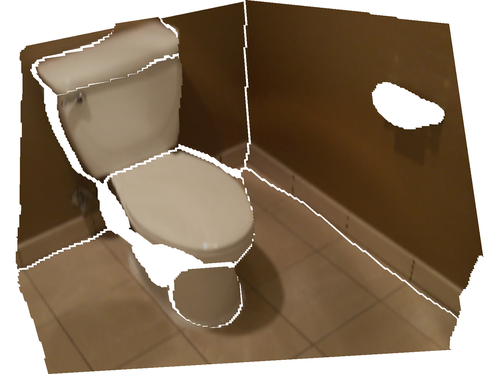}
		\tabularnewline
		& & & \smallskip{} 
		\tabularnewline		
		\includegraphics[width=0.25\textwidth]{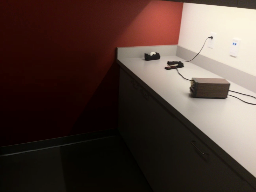}
		&\includegraphics[width=0.25\textwidth]{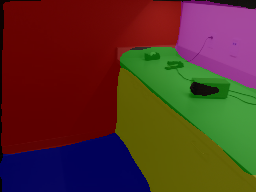}	
		&\includegraphics[width=0.25\textwidth]{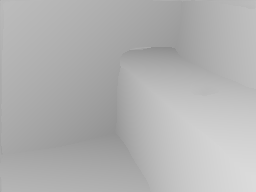}
		&\includegraphics[width=0.25\textwidth]{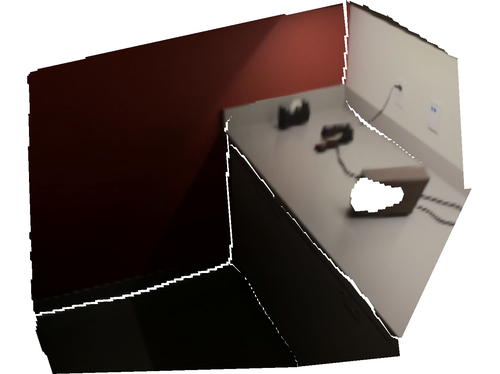}
		\tabularnewline
		& & & \smallskip{} 
		\tabularnewline
		{\small{}Input image}
		&{\small{}Plane instance segmentation} 
		&{\small{}Depth map}
		&{\small{}Piece-wise planar 3D model}
	\end{tabular}\hfill{}
	\caption{More piece-wise planar reconstruction results on the ScanNet dataset. In the plane instance segmentation results, black color indicates non-planar regions.\label{fig:scannet:more}}
\end{figure*}

\begin{figure*}[t]
	\centering
	\setlength{\tabcolsep}{0.1em}
	\renewcommand{\arraystretch}{0.5}
	\hfill{}\hspace*{-0.5em}
	\begin{tabular}{ccccc}
		\includegraphics[width=0.2\textwidth]{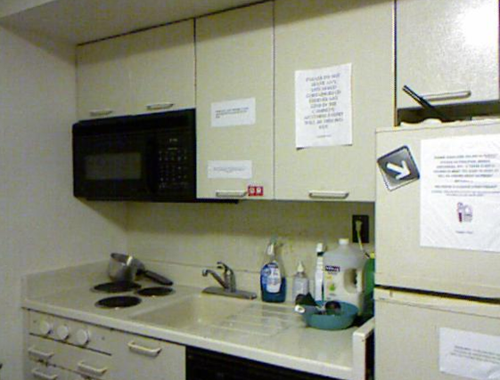}
		&\includegraphics[width=0.2\textwidth]{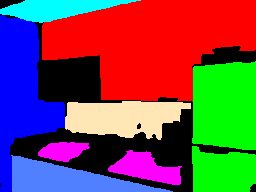}
		&\includegraphics[width=0.2\textwidth]{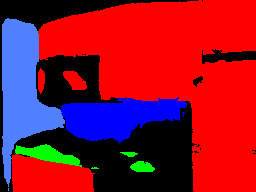}
		&\includegraphics[width=0.2\textwidth]{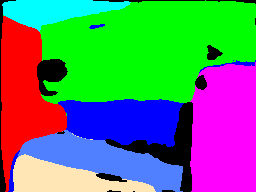}
		&\includegraphics[width=0.2\textwidth]{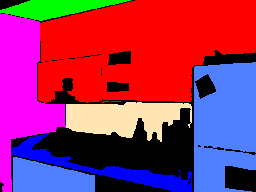}
		\tabularnewline
		\includegraphics[width=0.2\textwidth]{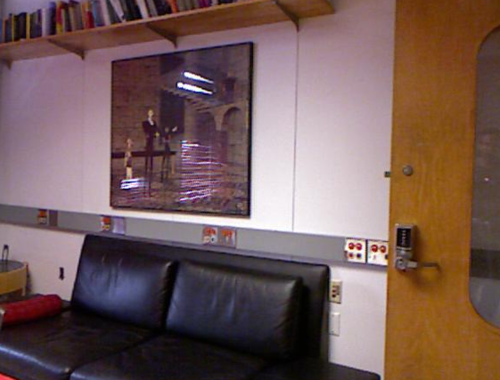}
		&\includegraphics[width=0.2\textwidth]{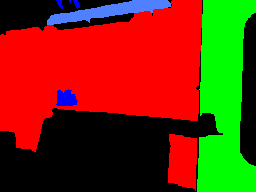}
		&\includegraphics[width=0.2\textwidth]{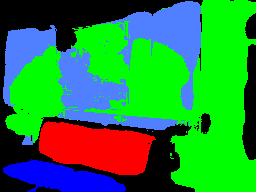}
		&\includegraphics[width=0.2\textwidth]{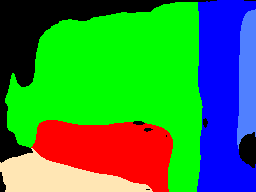}
		&\includegraphics[width=0.2\textwidth]{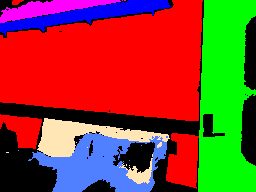}
		\tabularnewline
		\includegraphics[width=0.2\textwidth]{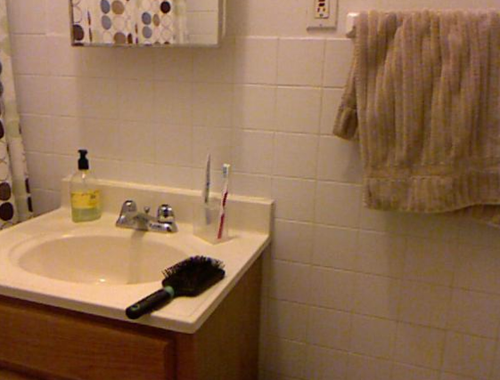}
		&\includegraphics[width=0.2\textwidth]{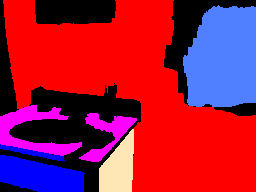}
		&\includegraphics[width=0.2\textwidth]{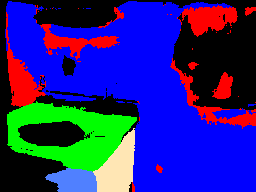}
		&\includegraphics[width=0.2\textwidth]{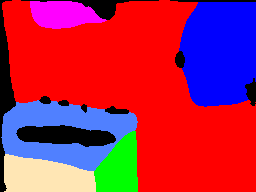}
		&\includegraphics[width=0.2\textwidth]{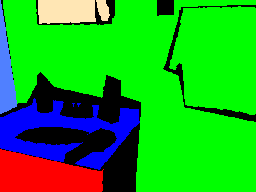}
		\tabularnewline
		\includegraphics[width=0.2\textwidth]{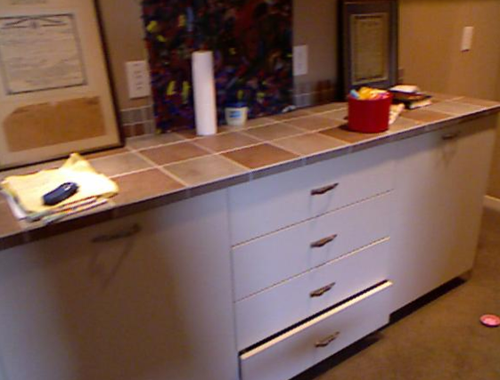}
		&\includegraphics[width=0.2\textwidth]{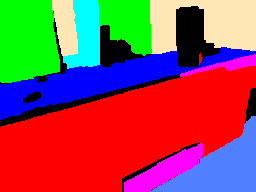}
		&\includegraphics[width=0.2\textwidth]{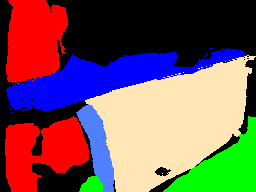}
		&\includegraphics[width=0.2\textwidth]{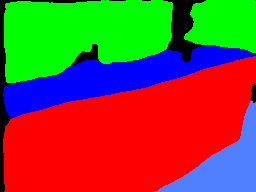}
		&\includegraphics[width=0.2\textwidth]{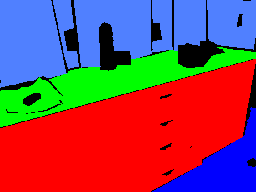}
		\tabularnewline
		\includegraphics[width=0.2\textwidth]{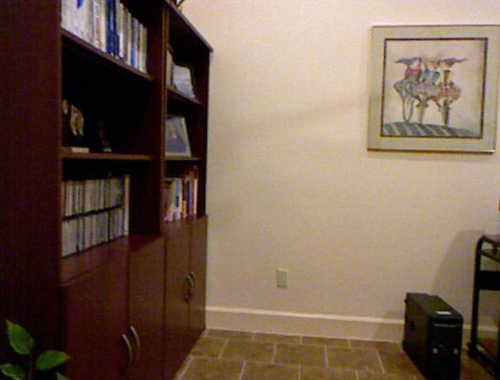}
		&\includegraphics[width=0.2\textwidth]{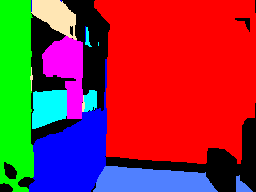}
		&\includegraphics[width=0.2\textwidth]{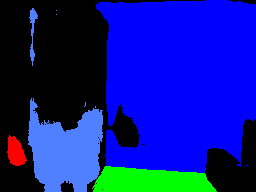}
		&\includegraphics[width=0.2\textwidth]{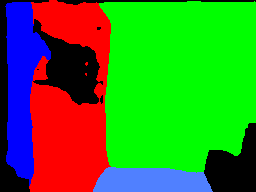}
		&\includegraphics[width=0.2\textwidth]{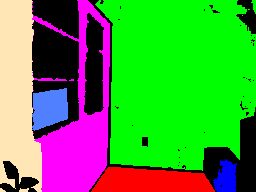}
		\tabularnewline
		\includegraphics[width=0.2\textwidth]{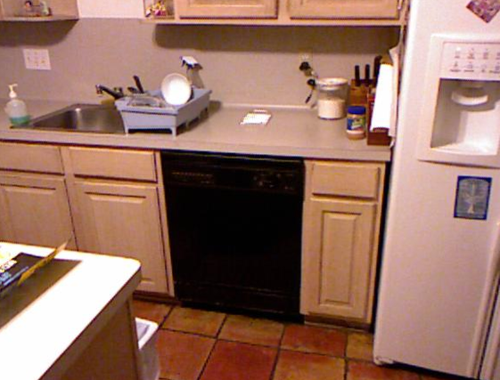}
		&\includegraphics[width=0.2\textwidth]{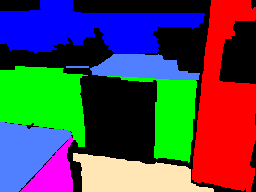} 
		&\includegraphics[width=0.2\textwidth]{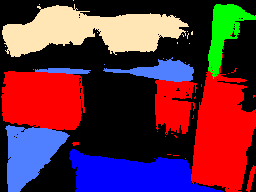}
		&\includegraphics[width=0.2\textwidth]{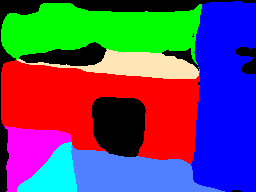}
		&\includegraphics[width=0.2\textwidth]{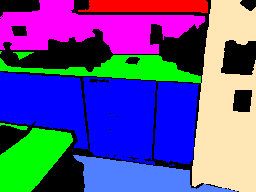}
		\tabularnewline
		\includegraphics[width=0.2\textwidth]{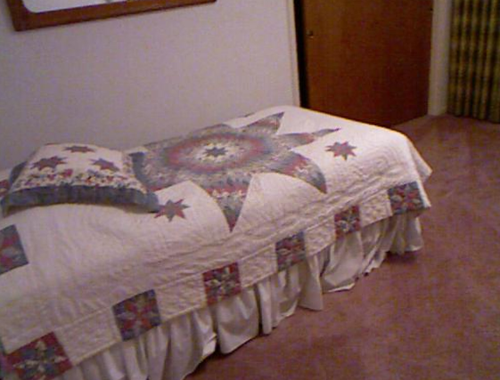}
		&\includegraphics[width=0.2\textwidth]{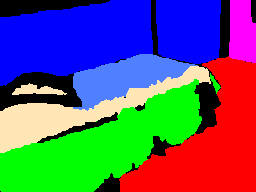} 
		&\includegraphics[width=0.2\textwidth]{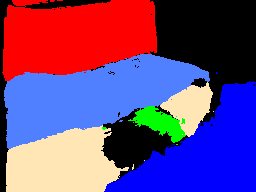}
		&\includegraphics[width=0.2\textwidth]{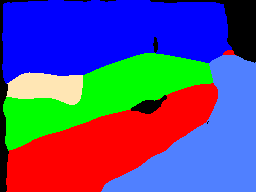}
		&\includegraphics[width=0.2\textwidth]{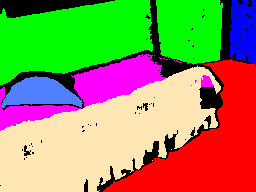}
		\tabularnewline
		{\small{}Input image}
		&{\small{}GT Depth + \cite{Silberman2012Indoor}}
		&{\small{}PlaneNet~\cite{Liu2018PlaneNet}} 
		&{\small{}Ours} 
		&{\small{}Ground truth}
	\end{tabular}\hfill{}
	\caption{More plane instance segmentation results on the NYUv2 dataset. Black color indicates non-planar regions.\label{fig:nyuv2}}
\end{figure*}

\PAR{Qualitative evaluation.}
Additional reconstruction results on the ScanNet dataset are shown in Figure~\ref{fig:scannet:more}. More qualitative comparisons against existing methods for plane instance segmentation on the NYUv2 dataset are shown in Figure~\ref{fig:nyuv2}. 

\begin{table}[t]
\centering
\caption{Comparison of depth prediction accuracy without fine-tuning on NYUv2 test set. Note that lower is better for top five rows, whereas higher is better for the bottom three rows.\label{tab:nyuv2:depth_wo}}
\vspace{-3mm}
\begin{tabular}{c|cc}
	Method &PlaneNet~\cite{Liu2018PlaneNet} &Ours
	\tabularnewline \hline\hline
	{Rel} &{\small{}0.238} &\textbf{\small{}0.219}
	\tabularnewline
	{Rel(sqr)} &{\small{}0.287} &\textbf{\small{}0.250}
	\tabularnewline
	{log$_{10}$} &{\small{}0.126} &\textbf{\small{}0.112}
	\tabularnewline
	{RMSE$_{iin}$} &{\small{}0.925} &\textbf{\small{}0.881}
	\tabularnewline
	{RMSE$_{\log}$} &{\small{}0.334} &\textbf{\small{}0.305}
	\tabularnewline \hline
	{1.25} &{\small{}49.1} &\textbf{\small{}53.3}
	\tabularnewline
	{1.25$^2$} &{\small{}79.0} &\textbf{\small{}84.5}
	\tabularnewline
	{1.25$^3$} &{\small{}91.9} &\textbf{\small{}95.1}
\end{tabular}
\vspace{-3mm}
\end{table}

\begin{table*}[t]
	\centering
	\setlength{\tabcolsep}{5pt}
	\renewcommand{\arraystretch}{1.1}
	\caption{Plane reconstruction accuracy comparisons on the ScanNet dataset.\label{tab:scannet:recall}}
	\subfloat[Plane recall versus depth difference.\label{tab:scannet:plane:depth}]{
		\begin{tabular}{cc|cccccccccccc}
			\multicolumn{2}{c|}{\footnotesize{}Depth threshold} 
			&{\small{}0.05} &{\small{}0.10} &{\small{}0.15} &{\small{}0.20} &{\small{}0.25} &{\small{}0.30} &{\small{}0.35} &{\small{}0.40} &{\small{}0.45} &{\small{}0.50} &{\small{}0.55} &{\small{}0.60}
			\tabularnewline
			\hline \hline
			\multirow{2}{*}{{\footnotesize{}GT Depth}}
			&{\footnotesize{}MWS~\cite{Furukawa2009Manhattan}}
			&{\small{}51.22} &{\small{}63.84} &{\small{}67.20} &{\small{}68.28} &{\small{}68.61} &{\small{}68.74} &{\small{}68.85} &{\small{}68.87} &{\small{}68.89} &{\small{}68.92} &{\small{}68.92} &{\small{}68.92} 
			\tabularnewline \cline{2-14}
			&{\footnotesize{}NYU-Toolbox~\cite{Silberman2012Indoor}}
			&{\small{}45.66} &{\small{}48.34} &{\small{}48.69} &{\small{}48.82} &{\small{}48.89} &{\small{}48.91} &{\small{}48.91} &{\small{}48.93} &{\small{}48.93} &{\small{}48.93} &{\small{}48.96} &{\small{}48.96}
			\tabularnewline \hline
			\multirow{4}{*}{{\footnotesize{}Inferred Depth}}
			&{\footnotesize{}MWS~\cite{Furukawa2009Manhattan}}
			&{\small{}1.69} &{\small{}5.32} &{\small{}8.84} &{\small{}11.67} &{\small{}14.40} &{\small{}16.97} &{\small{}18.71} &{\small{}20.47} &{\small{}21.68} &{\small{}23.06} &{\small{}24.09} &{\small{}25.13} 
			\tabularnewline \cline{2-14}
			&{\footnotesize{}NYU-Toolbox~\cite{Silberman2012Indoor}}
			&{\small{}3.14} &{\small{}9.21} &{\small{}13.26} &{\small{}16.93} &{\small{}19.63} &{\small{}21.41} &{\small{}22.69} &{\small{}23.48} &{\small{}24.18} &{\small{}25.04} &{\small{}25.50} &{\small{}25.85}
			\tabularnewline \cline{2-14}
			&{\footnotesize{}PlaneNet~\cite{Liu2018PlaneNet}}
			&{\small{}15.78} &{\small{}29.15} &{\small{}37.48} &{\small{}42.34} &{\small{}45.09} &{\small{}46.91} &{\small{}47.77} &{\small{}48.54} &{\small{}49.02} &{\small{}49.33} &{\small{}49.53} &{\small{}49.59}
			\tabularnewline \cline{2-14}
			&{\footnotesize{}Ours} 
			&{\small{}22.93} &{\small{}40.17} &{\small{}49.40} &{\small{}54.58} &{\small{}57.75} &{\small{}59.72} &{\small{}60.92} &{\small{}61.84} &{\small{}62.23} &{\small{}62.56} &{\small{}62.76} &{\small{}62.93}
		\end{tabular}
	}
	\hfill
	\subfloat[Pixel recall versus depth difference.\label{tab:scannet:pixel:depth}]{
		\begin{tabular}{cc|cccccccccccc}
			\multicolumn{2}{c|}{\footnotesize{}Depth threshold} 
			&{\small{}0.05} &{\small{}0.10} &{\small{}0.15} &{\small{}0.20} &{\small{}0.25} &{\small{}0.30} &{\small{}0.35} &{\small{}0.40} &{\small{}0.45} &{\small{}0.50} &{\small{}0.55} &{\small{}0.60}
			\tabularnewline
			\hline \hline
			\multirow{2}{*}{{\footnotesize{}GT Depth}}
			&{\footnotesize{}MWS~\cite{Furukawa2009Manhattan}}
			&{\small{}64.44} &{\small{}74.37} &{\small{}76.36} &{\small{}76.85} &{\small{}76.96} &{\small{}77.03} &{\small{}77.07} &{\small{}77.08} &{\small{}77.09} &{\small{}77.09} &{\small{}77.09} &{\small{}77.09} 
			\tabularnewline \cline{2-14}
			&{\footnotesize{}NYU-Toolbox~\cite{Silberman2012Indoor}}
			&{\small{}73.59} &{\small{}75.49} &{\small{}75.67} &{\small{}75.75} &{\small{}75.78} &{\small{}75.80} &{\small{}75.80} &{\small{}75.80} &{\small{}75.80} &{\small{}75.80} &{\small{}75.81} &{\small{}75.81}
			\tabularnewline \hline
			\multirow{4}{*}{{\footnotesize{}Inferred Depth}}
			&{\footnotesize{}MWS~\cite{Furukawa2009Manhattan}}
			&{\small{}2.40} &{\small{}8.02} &{\small{}13.70} &{\small{}18.06} &{\small{}22.42} &{\small{}26.22} &{\small{}28.65} &{\small{}31.13} &{\small{}32.99} &{\small{}35.14} &{\small{}36.82} &{\small{}38.09}
			\tabularnewline \cline{2-14}
			&{\footnotesize{}NYU-Toolbox~\cite{Silberman2012Indoor}}
			&{\small{}3.97} &{\small{}11.56} &{\small{}16.66} &{\small{}21.33} &{\small{}24.54} &{\small{}26.82} &{\small{}28.53} &{\small{}29.45} &{\small{}30.36} &{\small{}31.46} &{\small{}31.96} &{\small{}32.34}
			\tabularnewline \cline{2-14}
			&{\footnotesize{}PlaneNet~\cite{Liu2018PlaneNet}} 
			&{\small{}22.79} &{\small{}42.19} &{\small{}52.71} &{\small{}58.92} &{\small{}62.29} &{\small{}64.31} &{\small{}65.20} &{\small{}66.10} &{\small{}66.71} &{\small{}66.96} &{\small{}67.11} &{\small{}67.14}
			\tabularnewline \cline{2-14}
			&{\footnotesize{}Ours}
			&{\small{}30.59} &{\small{}51.88} &{\small{}62.83} &{\small{}68.54} &{\small{}72.13} &{\small{}74.28} &{\small{}75.38} &{\small{}76.57} &{\small{}77.08} &{\small{}77.35} &{\small{}77.54} &{\small{}77.86}
		\end{tabular}
	}
	\hfill
	\subfloat[Plane recall versus normal difference.\label{tab:scannet:plane:normal}]{
		\begin{tabular}{cc|cccccccccccc}
			\multicolumn{2}{c|}{\footnotesize{}Normal threshold} 
			&{\small{}2.5} &{\small{}5.0} &{\small{}7.5} &{\small{}10.0} &{\small{}12.5} &{\small{}15.0} &{\small{}17.5} &{\small{}20.0} &{\small{}22.5} &{\small{}25.0} &{\small{}27.5} &{\small{}30.0}
			\tabularnewline
			\hline \hline
			\multirow{2}{*}{{\footnotesize{}GT Normal}}
			&{\footnotesize{}MWS~\cite{Furukawa2009Manhattan}}
			&{\small{}39.19} &{\small{}54.03} &{\small{}58.93} &{\small{}61.23} &{\small{}62.69} &{\small{}64.22} &{\small{}64.90} &{\small{}65.58} &{\small{}66.15} &{\small{}66.61} &{\small{}67.13} &{\small{}67.29}
			\tabularnewline \cline{2-14}
			&{\footnotesize{}NYU-Toolbox~\cite{Silberman2012Indoor}}
			&{\small{}15.04} &{\small{}31.07} &{\small{}37.00} &{\small{}40.43} &{\small{}42.66} &{\small{}44.02} &{\small{}45.13} &{\small{}45.81} &{\small{}46.36} &{\small{}46.91} &{\small{}47.41} &{\small{}47.82}
			\tabularnewline \hline
			\multirow{4}{*}{{\footnotesize{}Inferred Normal}}
			&{\footnotesize{}MWS~\cite{Furukawa2009Manhattan}}
			&{\small{}1.73} &{\small{}05.79} &{\small{}10.04} &{\small{}13.71} &{\small{}16.23} &{\small{}18.22} &{\small{}19.48} &{\small{}20.71} &{\small{}21.69} &{\small{}22.50} &{\small{}23.25} &{\small{}23.60}
			\tabularnewline \cline{2-14}
			&{\footnotesize{}NYU-Toolbox~\cite{Silberman2012Indoor}}
			&{\small{}1.51} &{\small{}05.58} &{\small{}09.86} &{\small{}13.47} &{\small{}16.64} &{\small{}18.48} &{\small{}19.99} &{\small{}21.52} &{\small{}22.48} &{\small{}23.33} &{\small{}24.12} &{\small{}24.54}
			\tabularnewline \cline{2-14}
			&{\footnotesize{}PlaneNet~\cite{Liu2018PlaneNet}}
			&{\small{}12.49} &{\small{}29.70} &{\small{}40.21} &{\small{}44.92} &{\small{}46.77} &{\small{}47.71} &{\small{}48.44} &{\small{}48.83} &{\small{}49.09} &{\small{}49.20} &{\small{}49.31} &{\small{}49.38}
			\tabularnewline \cline{2-14}
			&{\footnotesize{}Ours} 
			&{\small{}20.05} &{\small{}42.66} &{\small{}51.85} &{\small{}55.92} &{\small{}58.34} &{\small{}59.52} &{\small{}60.35} &{\small{}60.75} &{\small{}61.23} &{\small{}61.64} &{\small{}61.84} &{\small{}61.93}
		\end{tabular}
	}
	\hfill
	\subfloat[Pixel recall versus normal difference.\label{tab:scannet:pixel:normal}]{
		\begin{tabular}{cc|cccccccccccc}
			\multicolumn{2}{c|}{\footnotesize{}Normal threshold} 
			&{\small{}2.5} &{\small{}5.0} &{\small{}7.5} &{\small{}10.0} &{\small{}12.5} &{\small{}15.0} &{\small{}17.5} &{\small{}20.0} &{\small{}22.5} &{\small{}25.0} &{\small{}27.5} &{\small{}30.0}
			\tabularnewline
			\hline \hline
			\multirow{2}{*}{{\footnotesize{}GT Normal}}
			&{\footnotesize{}MWS~\cite{Furukawa2009Manhattan}}
			&{\small{}56.21} &{\small{}70.53} &{\small{}73.49} &{\small{}74.47} &{\small{}75.12} &{\small{}75.66} &{\small{}75.88} &{\small{}76.04} &{\small{}76.28} &{\small{}76.41} &{\small{}76.55} &{\small{}76.59}
			\tabularnewline \cline{2-14}
			&{\footnotesize{}NYU-Toolbox~\cite{Silberman2012Indoor}}
			&{\small{}31.93} &{\small{}58.92} &{\small{}65.63} &{\small{}69.09} &{\small{}71.12} &{\small{}72.10} &{\small{}72.89} &{\small{}73.41} &{\small{}73.65} &{\small{}74.08} &{\small{}74.39} &{\small{}74.65}
			\tabularnewline \hline
			\multirow{4}{*}{{\footnotesize{}Inferred Normal}}
			&{\footnotesize{}MWS~\cite{Furukawa2009Manhattan}}
			&{\small{}2.58} &{\small{}8.51} &{\small{}15.08} &{\small{}20.16} &{\small{}24.51} &{\small{}27.78} &{\small{}29.63} &{\small{}31.96} &{\small{}33.65} &{\small{}34.99} &{\small{}36.37} &{\small{}37.03}
			\tabularnewline \cline{2-14}
			&{\footnotesize{}NYU-Toolbox~\cite{Silberman2012Indoor}}
			&{\small{}2.11} &{\small{}7.69} &{\small{}13.49} &{\small{}18.25} &{\small{}22.58} &{\small{}24.92} &{\small{}26.63} &{\small{}28.50} &{\small{}29.58} &{\small{}30.46} &{\small{}31.23} &{\small{}31.65}
			\tabularnewline \cline{2-14}
			&{\footnotesize{}PlaneNet~\cite{Liu2018PlaneNet}}
			&{\small{}19.68} &{\small{}43.78} &{\small{}57.55} &{\small{}63.36} &{\small{}65.27} &{\small{}66.03} &{\small{}66.64} &{\small{}66.99} &{\small{}67.16} &{\small{}67.20} &{\small{}67.26} &{\small{}67.29}
			\tabularnewline \cline{2-14}
			&{\footnotesize{}Ours}
			&{\small{}30.20} &{\small{}59.89} &{\small{}69.79} &{\small{}73.59} &{\small{}75.67} &{\small{}76.8} &{\small{}77.3} &{\small{}77.42} &{\small{}77.57} &{\small{}77.76} &{\small{}77.85} &{\small{}78.03}
		\end{tabular}
	}
\end{table*}
	
\end{appendices}

\end{document}